\documentclass{article} %
\usepackage[final]{colm2026_conference} %

\usepackage{microtype}
\usepackage{hyperref}
\usepackage{url}
\usepackage{booktabs}
\usepackage{xspace}
\usepackage{multirow}
\usepackage{graphicx}
\usepackage{wrapfig}
\usepackage[table]{xcolor}
\usepackage{amsmath}
\usepackage{makecell}
\usepackage{helvet}
\usepackage{xcolor}
\usepackage{tcolorbox}
\usepackage{fontawesome5}
\usepackage{fvextra}
\usepackage{floatrow}
\usepackage{subcaption}

\usepackage{lineno}
\usepackage{fontawesome5}
\usepackage{amsmath}

\definecolor{darkblue}{rgb}{0, 0, 0.5}
\hypersetup{colorlinks=true, citecolor=darkblue, linkcolor=darkblue, urlcolor=darkblue}

\newcommand{\frameworkName}{\mbox{\textsc{NormViz}}\xspace}
\newcommand{\frameworkNameWithEmoji}{\includegraphics[height=1em]{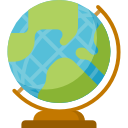}\mbox{\textsc{NormViz}}\xspace}

\newcommand{\benchmarkNameWithEmoji}{\mbox{\textsc{\frameworkNameWithEmoji-Bench}}\xspace}
\newcommand{\benchmarkName}{\mbox{\textsc{\frameworkName-Bench}}\xspace}
\newcommand{\trainName}{\mbox{\textsc{\frameworkName-Train}}\xspace}

\title{\frameworkNameWithEmoji: 
A Benchmark and Framework for Grounding Multimodal Reasoning in Global Cultures
}

\newcommand{\cmu}{$^\heartsuit$}
\newcommand{\ucla}{$^\clubsuit$}
\author{
Akhila Yerukola\cmu \quad
Fabrice Harel-Canada\ucla \quad
Simran Khanuja\cmu \quad
Abhinav Rao\cmu \AND
Ashima Suvarna\ucla \quad
Nanyun Peng\ucla \quad
Saadia Gabriel\ucla \quad
Maarten Sap\cmu \\ 
\cmu Carnegie Mellon University \quad \ucla University of California, Los Angeles \\ 
\hspace{\fill}\hspace{\fill}\faEnvelope~\texttt{\href{mailto:ayerukol@andrew.cmu.edu}{ayerukol@andrew.cmu.edu}}
}

\begin{document}

\ifcolmsubmission
\linenumbers
\fi

\maketitle

\begin{abstract}

AI systems are used worldwide, but they struggle to serve the needs of culturally diverse populations. Prior work on cultural understanding evaluates AI systems on text-only settings or on visual artifact recognition (e.g. foods, clothing). The ability to reason about visually observable behaviors through local social norms, which we call \emph{visual norm understanding}, remains unexamined. We introduce \benchmarkName, a high quality, human-validated benchmark of 3,272 contrastive image pairs (6,544 images) spanning 16 countries. Each pair varies only in the culturally relevant behavior (e.g., objects, attributes, spatial relations, and actions) that alters how each image is interpreted. Each image is labeled as conforming to, violating, or irrelevant to local social norms, and pair-level evaluation requires both images to be correctly classified, thereby preventing reliance on superficial visual shortcuts. Even the strongest VLMs, Gemini 3.0 Flash and Qwen2.5 VL 7B, succeed on only \textbf{25.3\%} and \textbf{23.0\%} of pairs, struggling most with identifying violating and culturally benign visual behaviors. Towards bridging this, we introduce \trainName, a training dataset of $64k$ images paired with explanations. Though absolute performance remains low ($<30\%$), finetuning on \trainName improves pair accuracy relatively by up to 86\% and 36\% for Qwen3-VL 4B and 8B respectively, showing a path forward to teach models to connect visual perception to cultural significance. Together, \benchmarkName and \trainName establish visual norm understanding as a challenging and consequential frontier for multimodal AI. We make our code and data publicly available.
{\footnote{Code and data available at: \url{https://github.com/Akhila-Yerukola/NormViz}}}
\looseness=-1

\end{abstract}

\section{Introduction}
\label{sec:introduction}

\begin{figure*}[t]
    \centering
    \includegraphics[width=0.9\columnwidth, trim={6em 0em 4em 4em}]{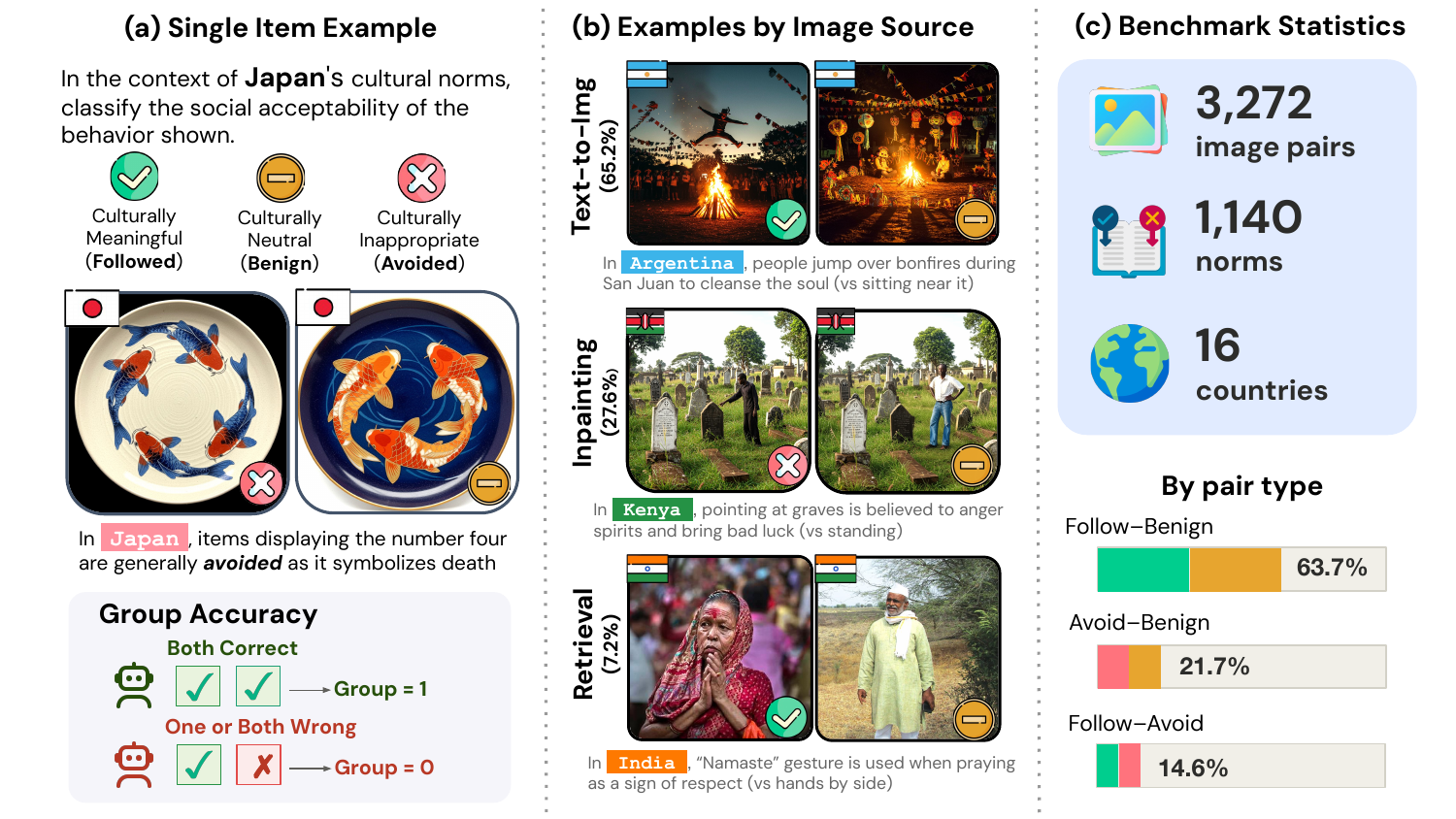}
    \caption{Overview of \benchmarkName for visual norm understanding, consisting of 3,272 contrastive image pairs across 16 countries that differ only in the culturally relevant behavior. (a) A contrastive pair example from Japan; group accuracy requires a model to correctly classify both images in a pair. (b) Images are sourced via Text-to-Image generation, inpainting, and Retrieval. (c) Benchmark statistics and pair type distribution.}
    
    \vspace{-1em}

    \label{fig:intro-figure}
\end{figure*}

As multimodal AI systems are increasingly deployed across culturally diverse settings, the ability to identify and reason about \textit{visually observable} social dynamics is a critical capability, which we refer to as \textbf{visual norm understanding} \citep{matsumoto2016cultural, knapp1972nonverbal, geertz2017interpretation}. Consider a celebratory event in Japan where a guest gifts a ceramic plate with four koi fish (Fig. \ref{fig:intro-figure} (a), left image). A Japanese native would recognize this as culturally offensive \citep[in Japanese the number 4, `shi', is homophonous with the word for `death';][]{simon1952some}; in contrast, a plate with three fish (Fig. \ref{fig:intro-figure} (a), right image) would be culturally benign --- neither meaningful nor offensive. A crucial step towards AI systems effectively serving users across global contexts is this ability to not just recognize \textit{what} is in an image, but also understand \textit{how} specific visual behaviors relate to local social and cultural norms.
\looseness=-1

Yet, current AI training and evaluation has largely ignored this critical ability.
Prior work on evaluating cultural understanding in AI has mostly operated in text-only settings, where cultural context is explicitly provided as input \citep{durmus2023towards, rao2025normad, kabra2023multi}. Efforts in multimodal understanding, meanwhile, have focused on \textit{cultural objects and artifacts} such as foods, clothing, and landmarks \citep{urailertprasert-etal-2024-sea, nayak-etal-2024-benchmarking, romero2024cvqa}. \textit{Culturally situated behaviors}, such as gifting practices, are hard to isolate in web-scale image data \citep{van2017cross,slingerland2020coding}, making it difficult to curate data that goes beyond artifact recognition to visual norm reasoning.
\looseness=-1

To fill this gap, we introduce \benchmarkNameWithEmoji, a high-quality, diverse, and human-validated benchmark of 3,272 contrastive image pairs (6,544 images) for evaluating visual norm understanding across 16 countries. 
We adopt contrastive pair evaluation since prior work show  models often exploit superficial shortcuts in single-image evaluative settings (e.g., easier to detect a cup `on' vs. `under' a table) whereas contrastive minimal-pair designs are more robust to such biases \citep{thrush2022winoground, kamath2023s}.
In our benchmark, each pair differs only in the culturally relevant behavior, requiring both visual perception and normative judgment.  
Given each image, models must determine if the depicted behavior conforms to (\textit{Follow}), violates (\textit{Avoid}), or is irrelevant (\textit{Benign}) to local social norms, and must classify both images correctly.
Visually detectable cultural norms are sourced from three human-validated text-based cultural understanding benchmarks \citep{chiu2024culturalbench, rao2025normad, yin2024safeworld}, transformed into images with benign counterparts through text-to-image generation, inpainting, and real image retrieval, with each pair rigorously validated by qualified annotators from the target country (Figure \ref{fig:intro-figure}).
\looseness=-1

Using \benchmarkName, we evaluate 21 vision-language models (VLMs) using \textit{group accuracy}, the percentage of contrastive pairs where both images are correctly classified. We find that even the strongest open and closed-source models, Qwen2.5 VL-7B and Gemini 3.0 Flash, achieve only \textbf{23.0\%} and \textbf{25.3\%} respectively. Models struggle most with behaviors that violate social norms or are culturally irrelevant, and perform worst on cultures from the Global South, including India, Indonesia, and Kenya. To disentangle sources of VLM failures, we provide models with (i) image descriptions alongside images, and (ii) descriptions in place of images. Performance improves in both settings, yet the strongest models achieve below $45.3\%$ in group accuracy. This gap suggests that cultural normative reasoning, not visual perception or vision-language misalignment, is the hardest barrier.
\looseness=-1

To improve visual norm understanding, we introduce \trainName, a large-scale synthetic training dataset of $64k$ images paired with explanations of the depicted behavior and its cultural significance. We source cultural norms from  CANDLE, a cultural knowledge graph \citep{nguyen2023extracting}, and generate images using Stable Diffusion 3 (SD3). Supervised finetuning Qwen3-VL models on \trainName yields meaningful improvements in group accuracy across model sizes: Qwen3-VL 4B improves group accuracy by 86\% (14.9\% to 27.7\%); Qwen3-VL 8B improves by 36\% (19.6\% to 26.7\%). With 64k training examples, both models \textbf{beat Gemini 3.0 Flash} (25.3\%). Despite these significant gains, absolute group accuracy remains below 30\%, establishing visual norm understanding as an open and consequential challenge that \benchmarkName and \trainName together help advance. \looseness=-1

\section{Related Work}
\label{sec:background}
\paragraph{AI Benchmarks for Cultural Understanding.}
Cultural competence in AI requires not just knowledge of diverse cultures, but also the ability to reason about cultural norms when interacting with or representing pluralistic needs \citep{deardorff2009sage, bhatt2024extrinsic}. Most AI benchmarks studying this have focused solely on text-based evaluation, assessing knowledge of cultural artifacts and facts \citep{chiu2024culturalbench, seth2024dosa, singh2025global}, cultural values and beliefs \citep{durmus2023towards, arora2023probing, cao2023assessing, ramezani2023knowledge}, and understanding social norms \citep{rao2025normad, shi2024culturebank}.

Multimodal benchmarks, in contrast, have primarily focused on recognition of cultural artifacts such as food, clothing, and landmarks \citep{liu2021visually, romero2024cvqa, nayak-etal-2024-benchmarking, vayani2025all, yue2024pangea, schneider-etal-2025-gimmick, satar2025seeing, kim2025world}, essentially evaluating object recognition rather than compositional reasoning. Cultural norms, however, depend on compositional reasoning of social behaviors: understanding not just what is present in a scene, but who performs an action, toward whom, and in what cultural setting \citep{knapp1972nonverbal, burgoon2011nonverbal}. VLMs are known to struggle with this kind of reasoning in general settings, exhibiting a bag-of-words failure mode and performing near chance on tasks requiring sensitivity to relational structure, object attributes, or actions \citep{yuksekgonuland, ma2023crepe, li2024naturalbench, thrush2022winoground, parcalabescu2022valse}. Yet whether this weakness extends to culturally situated visual settings remains unexplored \citep{yerukola2025mind, kim2025speaking, malakouti2025culture}. No existing work, to our knowledge, has systematically evaluated the visual norm understanding of culturally situated actions, gestures, and behaviors in VLMs. 
\vspace{-0.5em}

\paragraph{Data Bottleneck for Training Culturally Grounded Models.}
A key bottleneck for cultural alignment is the difficulty of mining high-quality data at scale that encodes not just cultural facts but also norms, values, and situated behaviors \citep{pawar2025survey}. In the text domain, recent work has addressed this through synthetic data such as LLM-generated cultural dialogues \citep{li2024culturepark, li2024culturellm, xu2025self}, multilingual critique data \citep{feng2025culfit}, and cross-cultural blended data \citep{yuan2024cultural}.
In the visual modality, the scarcity problem is more acute: cultural norms and social practices are difficult to isolate, label, and index in web data \citep{slingerland2020coding}, leaving multimodal approaches largely confined to artifact-centric image-QA pairs \citep{liu2025culturevlm, nyandwi2025grounding} or real images curated for cultural safety \citep{qiu2025multimodal}. No existing work has crossed the modality gap, using culturally grounded textual knowledge to generate visual training data covering social behaviors and norms that are systematically absent from naturally occurring image corpora.

\begin{figure*}[!t]
    \centering
    \includegraphics[width=1\columnwidth, trim={2em 0em 25em 4em}]{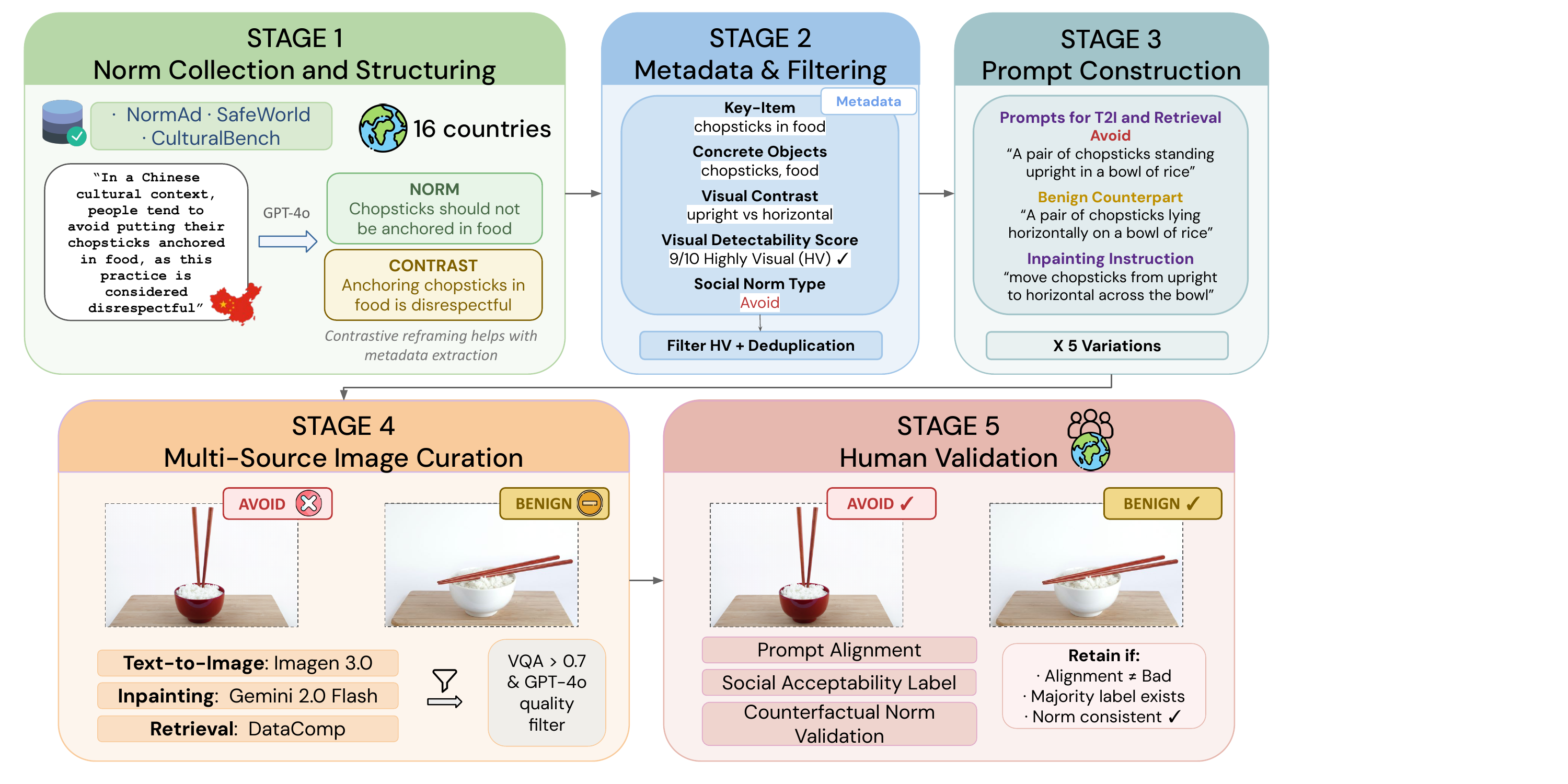}
    \caption{\benchmarkName construction pipeline across five stages: (1) collecting cultural norms from 3 text benchmarks and transforming them into norm-contrast pairs; (2) extracting metadata and social norm type, and then filtering for visual detectability; (3) constructing prompt descriptions of the contrasting pair, along with inpainting instructions; (4) curating images via text-to-image generation, inpainting, and retrieval, with initial quality filtering; and (5) human validation of the entire set from annotators from each of the 16 countries.
    }
    \label{fig:pipeline}
\end{figure*}

\section{\benchmarkNameWithEmoji Construction}
\label{sec:benchmark}

We curate \benchmarkName, a benchmark of contrastive image pairs for evaluating visual norm understanding in VLMs. Our pipeline (Fig. \ref{fig:pipeline}) transforms cultural norm descriptions into image pairs through five stages: (1) norm extraction and structuring (\S\ref{ssec:norm-extraction}), (2) metadata extraction and filtering for visual detectability (\S\ref{ssec:metadata}), (3) prompt construction (\S\ref{ssec:prompt_construction}), (4) multi-source image curation (\S\ref{ssec:image_curation}), and (5) human validation (\S\ref{ssec:human_validation}). 

\paragraph{Task Design Motivation.} \label{para:motivation} Prior work on (general) visual compositional reasoning has shown that single-image evaluation design fails to expose the visual and linguistic shortcuts that models exploit \citep{yuksekgonuland, ma2023crepe, li2024naturalbench, hsieh2023sugarcrepe}. Thus, we adopt a contrastive minimal-pair evaluation setup \citep{gardner2020evaluating, parcalabescu2022valse, thrush2022winoground}, where the two images differ only in culturally relevant behavior (e.g, number of items in Figure \ref{fig:intro-figure} (a)), and models must correctly classify both. A model that can correctly classify only one of the images might be relying on superficial features, rather than truly understanding the cultural norm.

\paragraph{Task Formulation.} 
Each image in these pairs depicts nonverbal visual behaviors (gestures, object attributes, spatial arrangements, or actions). Given an image and its country context, models must independently classify the depicted behavior as Culturally Significant (\textit{Follow}), Culturally Inappropriate (\textit{Avoid}), or Culturally Neutral (\textit{Benign}). We evaluate performance through two metrics: (1) pairwise classification via \textbf{group accuracy}, where both images in a pair must be correctly classified, (2) \textbf{single-image classification F1} performance.

\subsection{Stage 1: Norm Collection and Structuring}
\label{ssec:norm-extraction}
\textbf{Norm Collection.}
We source descriptions of cultural norms from three text benchmarks: CulturalBench \citep{chiu2024culturalbench}, NormAd \citep{rao2025normad}, and SafeWorld \citep{yin2024safeworld}, selected for their broad cultural coverage and extensive human validation. `Culture' is multifaceted, and we focus on several visual nuances covered across these benchmarks, such as social etiquette, religious practices, celebrations, and daily life. We target 16 countries prioritizing the Global South and based on data availability: Argentina, Brazil, China, Egypt, France, India, Indonesia, Italy, Japan, Kenya, Nigeria, Peru, South Korea, Turkey, United States, and Vietnam (spans 10 UN geoscheme regions). While culture transcends national borders, countries serve as a meaningful proxy \citep{adilazuarda2024measuring}, reflecting shared history and lived experience, and providing a practical starting point for geo-tagging data. 

\textbf{Norm Structuring.} 
We transform each cultural description into a \textit{norm--contrast} pair using GPT-4o,\footnote{We qualitatively validated GPT-4o outputs at every stage of our pipeline, and the final test set undergoes comprehensive human validation (\S\ref{ssec:human_validation})} extracting a behavioral norm (e.g., ``chopsticks should not be anchored in food'') and rephrasing it as a contrastive statement that preserves its meaning (e.g., ``anchoring chopsticks in food is considered disrespectful''). We found that visual metadata extraction is significantly more reliable when using norm–contrast pairs rather than norms alone.

\subsection{Stage 2: Metadata and Filtering}
\label{ssec:metadata}

\textbf{Metadata Extraction.} Not all cultural norms can be meaningfully depicted in images. For example, norms about attitudes (e.g., ``remain humble'') or verbal conventions (e.g., ``honorific titles for elders'') lack concrete visual anchors. To identify visually observable norms, we use GPT-4o to extract structured metadata for each \textit{norm--contrast} pair, spanning \textit{key items}, \textit{concrete items}, \textit{visual contrast}, among others (see Appendix \ref{app:ssec:metadata} for more details). We additionally classify each norm as \textit{Follow} or \textit{Avoid} based on its social norm type.

\textbf{Visual Detectability Filtering and Deduplication.}  Using the extracted metadata, we prompt GPT-4o to score each norm on visual detectability (0--10) based on clarity of representation and ease of distinguishability, retaining norms scoring $\geq$ 7/10 as Highly Visual (HV). 
For example, ``the \textit{bride}'s family gifts a Japanese tea set" (Low Visual; LV) is filtered out as it's hard to visually determine which family gave the gift. We then embed the filtered \textit{key items} and apply K-means clustering to merge semantically equivalent norms across the three source benchmarks, reducing the norm set by two-fold.

\subsection{Stage 3: Prompt Construction}
\label{ssec:prompt_construction}

For each norm classified as either \textit{Follow} or \textit{Avoid}, we use GPT-4o to generate three prompt types conditioned on the extracted metadata: (1) \textbf{\textit{Follow/Avoid} image descriptions}: detailed scene descriptions depicting the relevant social norm behavior; (2) \textbf{Benign counterfactual descriptions}: scenes retaining the key item in a culturally neutral configuration---neither meaningful nor offensive (e.g., ``A person holding chopsticks while reaching for food'').
We construct \textit{Benign} counterfactuals for both the norms to \textit{Follow} (\textit{Benign\textsubscript{Follow}}) and \textit{Avoid} (\textit{Benign\textsubscript{Avoid}}); (3) \textbf{Inpainting instructions}: minimal editing directives to transform a \textit{Follow/Avoid} image into its Benign counterpart (e.g., ``Change the red envelope to a yellow envelope''). We generate 5 variations of prompts  per category to increase scene diversity.

\subsection{Stage 4: Multi-Source Image Curation}
\label{ssec:image_curation}

For each norm, we curate both synthetic and real images to ensure diversity in visual style. With a growing prevalence of AI-generated content, there is a need to detect cultural inconsistencies in generated images, as well as to understand how social norms manifest in real images within target cultures.

\textbf{Text-to-Image Generation.} We first generate images for both \textit{Avoid/Follow} and \textit{Benign} behaviors independently. We select \texttt{Imagen 3.0} for its high-quality and culturally competent image generation \citep{kannen2024beyond}. We generate 2 images per prompt variation.

\textbf{Inpainting.} To create  challenging image pairs with minimal variation, we apply inpainting instructions only to the generated \textit{Follow/Avoid} images (e.g., change a red envelope to yellow). 
We select Gemini 2.0 Flash \citep{comanici2025gemini} for its strong inpainting capabilities that preserve image coherence while making precise edits to produce \textit{Benign} counterparts.

\textbf{Real Image Retrieval.} Generated images may not capture how these behaviors naturally appear in the target country, so we retrieve from country specific image databases instead. We use the \textit{Avoid, Follow, Benign} prompts to query country-specific subsets of DataComp \citep{gadre2023datacomp}, and retrieve up to 20 candidates per prompt.

\paragraph{Quality Filtering.} Prompt-image alignment is difficult to control at scale, particularly when models misinterpret or refuse culturally specific prompts \citep{yerukola2025mind}. Before human validation (§\ref{ssec:human_validation}), we apply two types of automatic filtering: (1) VQA-based scoring \citep{lin2024evaluating}, retaining images with semantic alignment above 0.7 (scale 0--1), and (2) GPT-4o-based evaluation verifying that key objects, attributes, and spatial relationships are correctly depicted, classifying each image as good, partial, or bad. Only good images are retained.
\looseness=-1

\subsection{Human Validation}
\label{ssec:human_validation}

To ensure the ecological validity of our benchmark, we recruited cultural in-group annotators from each of the 16 countries via Prolific (\url{https://www.prolific.com/}). Annotators were prescreened using a pre-qualification test to ensure annotation quality. We obtain 3 annotations per image pair. For each pair, annotators perform three tasks: (1) rate the \textbf{prompt-image alignment} for each image (good, partial, or bad); (2) independently \textbf{classify each image} as culturally appropriate, culturally inappropriate, or culturally benign; and (3) verify that \textbf{both images depict the same cultural norm} (\textit{counterfactual validation}). To ensure benchmark quality, we retain only pairs where (1) neither image is rated as a bad prompt-image match, (2) annotators reach majority agreement on cultural alignment labels, and (3) norm consistency is confirmed. Since labels are assigned freely, the resulting pairs are not constrained to the intended \textit{Follow/Avoid--Benign} structure, resulting in \textit{Avoid--Benign}, \textit{Follow--Benign}, and \textit{Follow--Avoid} pairs, allowing annotators to surface and correct mislabeled or culturally ambiguous cases. We filter out pairs with the same label. 

After our initial quality filtering (\S\ref{ssec:image_curation}), we obtained 8,456 candidate image pairs for annotation. We recruited 212 annotators across 16 countries (113 male, 98 female, 1 undisclosed), with each annotator evaluating 70-150 pairs (varied due to  availability). After filtering spam annotations, we retained 21,743 annotations. Of these, 38.6\% passed rigorous quality checks, yielding the final benchmark of 3,272 image pairs across 16 countries. This rigorous filtering process also specifically guards against any potential AI generator-introduced biases being passed through to the benchmark.  Please find details on the prescreening, annotation scheme, IRB approval and fair pay in Appendix \ref{app:annotation_details}.

\subsection{\benchmarkName Characteristics}
\label{sec:dataset-characteristics}

Our final benchmark comprises \textbf{3,272 contrastive image pairs} (\textbf{6,544 images}; Table \ref{app:tab:dataset_stats})  spanning \textbf{16 countries} from three image sources: \textit{text-to-image generation} (65.2\%), \textit{inpainting } (27.6\%), and \textit{natural image retrieval} (7.2\%), with three pair types: \textit{Follow--Benign} (63.7\%), \textit{Avoid--Benign} (21.7\%), and \textit{Follow--Avoid} (14.6\%). Notably, the retrieval subset of 7.2\% reflects how rarely these norms manifest in naturally occurring real images, and their retrieval is non-trivial. \benchmarkName covers a semantically rich range of cultural nuances---including public conduct, greetings and gestures, festivals, symbolic objects and practices, gift-giving and dining customs---and requires diverse visual reasoning over contrasting object states, actions, presence/absence, and gesture differences.

The per-country distribution in \benchmarkName is uneven (Figure \ref{app:fig:per_country_sources}; Vietnam: 410 pairs, China: 344, vs.\ United States: 92), partly due to the country coverage decisions inherited from our seed benchmarks (see \S\ref{ssec:norm-extraction}), but also reflects a well-established distinction between high-context and low-context cultures in cross-cultural communication theory \citep{hall1976beyond, hall2000context}. High-context cultures (e.g., Japan, China) embed meaning in nonverbal cues, body language, and situational context, which are precisely the visually observable norms our benchmark targets, whereas low-context cultures (e.g., the United States) rely more on explicit verbal communication and thus exhibit fewer visually distinctive, detectable nonverbal norms. Please refer to Appendix \ref{app:sec:data_statistics} and \ref{app:benchmark_construction} for full details on benchmark construction, prompts used, and benchmark characteristics.

\section{Benchmarking: Setup, Results, and Analysis}
\label{sec:results}
\paragraph{Setup and Metrics.} We evaluate 21 VLMs covering proprietary and open-weight systems. We prompt models with each image and ask them to classify its social acceptability within a specified cultural context: \textit{"In the context of \{country\}'s cultural and social norms, classify the social acceptability of the behavior shown in this image."}. Models choose from three options: \textit{(A) Culturally Meaningful/Significant, (B) Culturally Inappropriate/Offensive, or (C) Culturally Benign/Neutral}. We set temperature to 0.0 for all experiments. We report two metrics: (1) \textbf{Group accuracy} for pairwise classification, and (2) \textbf{macro F1} for single-image classification.

\definecolor{ablationgray}{gray}{0.94}
\definecolor{textgray}{gray}{0}

\newcolumntype{C}{>{\centering\arraybackslash}c}

\newcommand{\ab}[1]{#1}
\newcommand{\abavg}[1]{\textbf{ #1}}

\begin{table*}[!htbp]
\centering
\setlength{\tabcolsep}{3pt}
\renewcommand{\arraystretch}{1.05}
\resizebox{\textwidth}{!}{

\begin{tabular}{@{}
  r @{\hspace{2.5pt}}
  l
  C C C
  @{\hspace{3pt}} !{\color{gray!50}\vrule} @{\hspace{3pt}}
  C C C C C C C C C
  @{\hspace{3pt}} !{\color{gray!50}\vrule} @{\hspace{3pt}}
  C
  @{}}

\toprule

 & & \multicolumn{3}{c}{\textit{Closed source}}
 & \multicolumn{9}{c}{\textit{Open source}}
 & \\

\cmidrule(lr){3-5} \cmidrule(lr){6-14}
& \textbf{Condition}
  & \makecell{Gemini 3\\flash} & \makecell{GPT\\5.2} & \makecell{GPT\\4o} & \makecell{Qwen 2.5\\7B} & \makecell{LLaVA\\OV 7B} & \makecell{Qwen 3\\8B} & \makecell{Qwen 2.5\\3B} & \makecell{GLM-4V\\9B} & \makecell{Phi-3V\\4.2B} & \makecell{Aya\\Vis. 8B} & \makecell{Qwen 3\\4B} & \makecell{DS-vl2\\ 1B}
  & \textbf{Avg} \\
\midrule
\multicolumn{15}{@{}l}{\textit{\small Group Accuracy (\%)}} \\[1.5pt]

& \benchmarkName
  & \textbf{25.3} & \underline{22.7} & 16.3 & \textbf{23.0} & \underline{19.8} & 19.6 & 18.3 & 16.7 & 16.1 & 15.4 & 14.9 & 13.8 & \textbf{18.5} \\

\multirow{2}{*}{\rotatebox{90}{\textcolor{textgray}{\scriptsize\textit{oracle}}}}
  & \cellcolor{ablationgray}\ab{\small + Image Descriptions}
  & \cellcolor{ablationgray}\ab{32.7} & \cellcolor{ablationgray}\ab{37.8} & \cellcolor{ablationgray}\ab{23.5} & \cellcolor{ablationgray}\ab{30.8} & \cellcolor{ablationgray}\ab{27.3} & \cellcolor{ablationgray}\ab{28.1} & \cellcolor{ablationgray}\ab{27.5} & \cellcolor{ablationgray}\ab{23.1} & \cellcolor{ablationgray}\ab{29.9} & \cellcolor{ablationgray}\ab{26.7} & \cellcolor{ablationgray}\ab{23.0} & \cellcolor{ablationgray}\ab{19.7} & \cellcolor{ablationgray}\abavg{27.5} \\
  & \cellcolor{ablationgray}\ab{\small + Image Desc. (w/o img)}
  & \cellcolor{ablationgray}\ab{40.8} & \cellcolor{ablationgray}\ab{45.3} & \cellcolor{ablationgray}\ab{37.2} & \cellcolor{ablationgray}\ab{32.7} & \cellcolor{ablationgray}\ab{34.3} & \cellcolor{ablationgray}\ab{40.2} & \cellcolor{ablationgray}\ab{29.7} & \cellcolor{ablationgray}\ab{31.0} & \cellcolor{ablationgray}\ab{35.8} & \cellcolor{ablationgray}\ab{28.7} & \cellcolor{ablationgray}\ab{30.7} & \cellcolor{ablationgray}\ab{2.7} & \cellcolor{ablationgray}\abavg{32.4} \\

\midrule
\multicolumn{15}{@{}l}{\textit{\small Macro F1}} \\[1.5pt]

& \benchmarkName
  & \textbf{55.8} & \underline{45.6} & 41.5 & 44.2 & \underline{44.3} & 43.6 & \textbf{44.6} & 42.5 & 40.2 & 37.3 & 37.7 & 40.6 & \textbf{43.2} \\

\multirow{2}{*}{\rotatebox{90}{\textcolor{textgray}{\scriptsize\textit{oracle}}}}
  & \cellcolor{ablationgray}\ab{\small + Image Descriptions}
  & \cellcolor{ablationgray}\ab{63.6} & \cellcolor{ablationgray}\ab{60.8} & \cellcolor{ablationgray}\ab{50.5} & \cellcolor{ablationgray}\ab{48.9} & \cellcolor{ablationgray}\ab{48.5} & \cellcolor{ablationgray}\ab{50.0} & \cellcolor{ablationgray}\ab{52.2} & \cellcolor{ablationgray}\ab{47.4} & \cellcolor{ablationgray}\ab{49.9} & \cellcolor{ablationgray}\ab{44.3} & \cellcolor{ablationgray}\ab{43.7} & \cellcolor{ablationgray}\ab{45.4} & \cellcolor{ablationgray}\abavg{50.4} \\
  & \cellcolor{ablationgray}\ab{\small + Image Desc. (w/o img)}
  & \cellcolor{ablationgray}\ab{68.0} & \cellcolor{ablationgray}\ab{66.5} & \cellcolor{ablationgray}\ab{62.8} & \cellcolor{ablationgray}\ab{53.2} & \cellcolor{ablationgray}\ab{53.6} & \cellcolor{ablationgray}\ab{60.2} & \cellcolor{ablationgray}\ab{56.0} & \cellcolor{ablationgray}\ab{53.9} & \cellcolor{ablationgray}\ab{56.4} & \cellcolor{ablationgray}\ab{48.5} & \cellcolor{ablationgray}\ab{52.9} & \cellcolor{ablationgray}\ab{26.8} & \cellcolor{ablationgray}\abavg{54.9} \\

\bottomrule
\\
\end{tabular}%

}

\caption{\normalfont
  Performance across the top 12 evaluated models, ordered by group accuracy within closed- and open- source groups. \textbf{Best} and \underline{second-best} model are highlighted. \textbf{Group accuracy for all models is critically low}. We run two oracle ablations: (1) \textit{+ Image Descriptions} - image with text description; (2)  \textit{+ Image Desc. (w/o img)} - replace input image entirely with text description. We find that performance generally improves in both settings, indicating cultural visual perception and vision-language alignment are bottlenecks. However, the best model under these settings achieves 45.3\% group accuracy (GPT 5.2), suggesting that cultural normative reasoning is the biggest barrier. Full results for all 21 models in App. \ref{app:sec:benchmarking_results}.
}
\label{tab:overall_results_tab}
\end{table*}

\paragraph{Main results.}
Current VLMs struggle significantly with visually grounded cultural reasoning. Across all models, \textbf{group accuracy is strikingly low}, with the best-performing models Gemini 3.0 flash, GPT 5.2, Qwen2.5 VL 7B, and LLaVA OneVision-7B  achieving only 25.3--19.8\% (see Table~\ref{tab:overall_results_tab}; full results in Table~\ref{app:tab:all_models}). \textbf{Macro F1 scores for single-image classification are also weak}, ranging from 55.8 to 44.2 across these top models. This gap reveals a fundamental limitation: since contrastive pairs differ only in a culturally relevant visual detail, getting both images right is necessary for assessing visual norm understanding. Relatively stronger F1 but strikingly low group accuracy suggests models rely on shallow cultural associations rather than the subtle visual distinctions that separate these categories.
Performance further varies substantially across regions: \textbf{Sub-Saharan African, Southern Asian, and Southeast Asian regions show the lowest group accuracy}, while Western European and South American countries (particularly Argentina and Brazil) perform comparatively better (Appendix  \ref{app:ssec:per_country}).  Finally, while larger models tend to perform slightly better, \textbf{scaling effects are inconsistent across model families}. Refer to Appendix~\ref{app:sec:benchmarking_results} for more details.
\looseness=-1

\paragraph{Three Barriers: Visual Perception, Cultural Vision-Language Alignment, and Cultural Normative Reasoning}

To probe whether failures stem from visual perception, vision-language misalignment, or cultural reasoning, we conduct two oracle ablations using \benchmarkName (rows 2, 3 in Table \ref{tab:overall_results_tab}). First, we provide models with ground-truth image descriptions alongside the image (\textit{+ Image Descriptions}).\footnote{Image descriptions are the exact prompts used for image generation and retrieval  (e.g., ``A pair of chopsticks standing upright in a bowl of rice'')} Second, we remove the image entirely, providing only the image description (\textit{+ Image Desc. w/o img}). We note that these ablations do not perfectly isolate each factor and are best interpreted as generous upper bounds rather than clean separations. Nonetheless, the consistent significant performance improvements are indicative of several barriers. 

The first ablation yields consistent improvements (avg. +9.0 group accuracy, +7.3 F1; Wilcoxon signed-rank across models, $p<0.001$ for both metrics), indicating that visual perception is a significant bottleneck with VLMs struggling to correctly interpret what is depicted in culturally-laden images. Yet absolute performance achieved remains critically low (avg. 27.5\% group accuracy, 50.4 F1), suggesting that perception alone does not explain the failures. Strikingly, removing the image entirely improves performance further (avg. 32.4\% group accuracy, 54.9 F1; Wilcoxon signed-rank, $p<0.05$ for both metrics). 
This points to a vision-language misalignment between encoding culturally-specific image content and its textual description. Modern VLMs are trained to align visual content with general linguistic descriptions \citep{yu2022coca, tschannen2025siglip, bai2025qwen3}, but perhaps \textit{cultural} alignment is sparse in that training signal, causing the two signals to conflict rather than reinforce each other---turning the image into a source of noise rather than information.
Yet even under this most favorable condition, absolute performance remains far from perfect. This points to a second, deeper barrier: models lack the cultural knowledge to make normative judgments about the behaviors described, regardless of how those behaviors are conveyed. Visual perception and cultural VL alignment are meaningful bottlenecks, but the ability to reason about the cultural appropriateness remains the harder barrier to bridge.

\begin{figure}[!t]
    \centering
    \includegraphics[width=\columnwidth, trim={0 0em 2em 2em}]{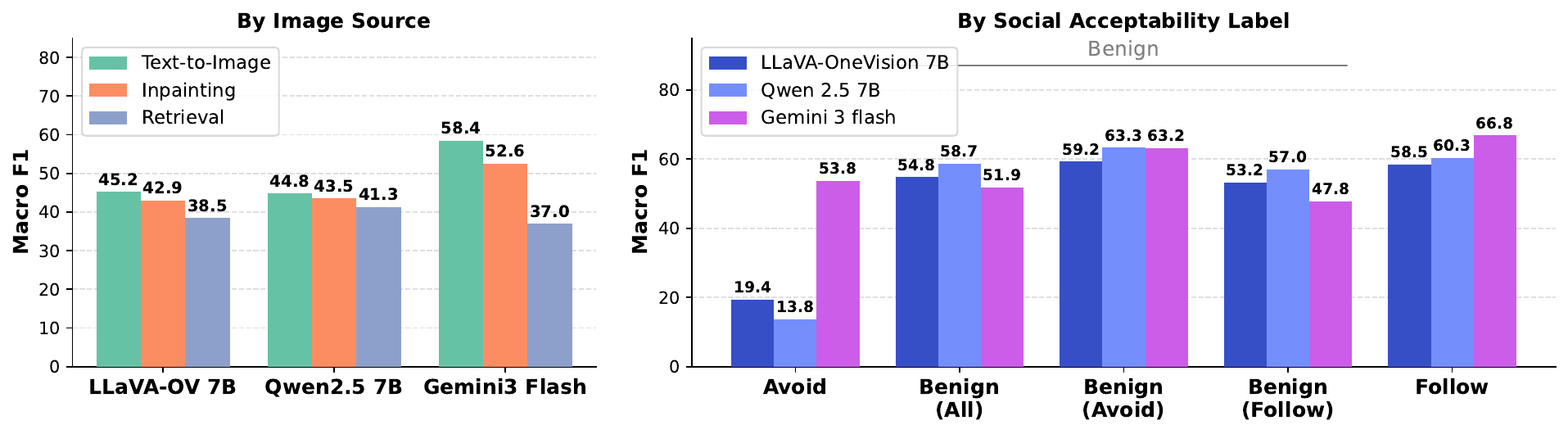}
    \caption{Single-image F1 performance of the top three models: (a) across data sources and (b) across social acceptability labels. Naturally retrieved images are the most challenging, followed by inpainted images, with independently generated images being the easiest. Detecting offensive content is the most difficult task, while identifying the benign counterparts of offensive behavior is easier than detecting offensive content.}
    \label{fig:performance_by_label_source}
     \vspace{-1em}
\end{figure}

\textbf{Performance by Data Source.}
Figure \ref{fig:performance_by_label_source} shows the best closed source model (Gemini 3 Flash) and top 2 open source models (Qwen 2.5 VL 7b, LLaVa-OneVision 7b). We find that retrieval-based images are the hardest (F1 drops 2--21 points), followed by inpainted images, while independently generated images are easiest (Figure \ref{fig:performance_by_label_source} (a)). Natural retrieval-based images likely contain more visual variability and noise than synthetic alternatives. For inpainting, the minimal differences between image pairs likely make it harder to detect changes in cultural behaviors, whereas independently generated pairs tend to exhibit more pronounced differences, making it easier to identify them.
\looseness=-1

\textbf{Performance by Social Acceptability Labels.}
As seen in Figure \ref{fig:performance_by_label_source} (b), models are considerably better at detecting culturally significant behavior (\textit{Follow}, 58--67 F1) than culturally offensive behavior (\textit{Avoid}, 14--54 F1), with Qwen 2.5 7B exhibiting the most extreme disparity (58.7 vs.\ 13.8). For \textit{Benign} images, models more reliably identify the benign counterpart of an \textit{Avoid} norm than that of a \textit{Follow} norm (e.g., Qwen2.5 7B: 63.3 vs.\ 57.0), suggesting that, to models, the \textit{visual contrast of benign situations with offensive behavior is more salient} than the contrast with culturally significant behavior.

\section{Improving Cultural Visual Reasoning}
\label{sec:finetuning}

Our ablations reveal three sources of failure in VLMs: difficulty perceiving cultural content, vision-language misalignment, and limited cultural knowledge and normative reasoning capabilities. We hypothesize these gaps stem partly from training data limitations: current VLMs rarely encounter culturally grounded examples that connect visual content to cultural significance \citep{van2017cross}. However, mining high-quality data at scale for improving cultural competence remains an open and hard challenge \citep{slingerland2020coding}. To test whether targeted finetuning on synthetically generated data can help fill these gaps, we construct \trainName and finetune Qwen3-VL models.

\subsection{\trainName Construction}
\label{ssec:train-data}
We source descriptions of cultural norms from CANDLE \citep{nguyen2023extracting}, a large-scale broad-coverage cultural knowledge graph extracted from C4 web crawls \citep{raffel2020exploring}. We use GPT-4o to filter out norms appearing in \benchmarkName's evaluation set to prevent data leakage. Then we apply the same pipeline as \S\ref{sec:benchmark}, limiting to text-to-image generation to prioritize data scale over source diversity. We use Stable Diffusion 3 (SD3) \citep{esser2024scaling}, a strong open-source T2I model, to avoid closed-source API costs and forgo human validation. Hence, the resulting training set thus consists of cultural norms explicitly disjoint from \benchmarkName, and further differs from it in image source, using SD3 rather than Imagen 3.0 (and Gemini 2.0 Flash for inpainting) used in \benchmarkName. Although the resulting training dataset may contain inconsistencies, noisy training data remains beneficial at scale \citep{karpukhin2019training, nguyen2024multilingual, zhai2023sigmoid, goyal2024scaling}, and no such large-scale culturally-grounded training resource otherwise exists.  To further mitigate quality concerns, we aggressively filter image-text pairs using VQA scoring (as in \S\ref{ssec:image_curation}).
Please refer to Appendix \ref{app:ssec:train_construction_details} for additional details on decontamination and data-quality efforts.

\subsection{Supervised Finetuning (SFT)}
\label{sec:finetune-setup}
Training data is formatted as VQA instances mirroring the evaluation setup in \S\ref{sec:results}. Each instance consists of an image, the question\footnote{"In the context of \{country\}'s cultural and social norms, classify the social acceptability of the behavior shown in this image."} with answer options (representing \textit{Follow}, \textit{Avoid}, or \textit{Benign}), and a structured explanation generated by GPT-4o. The explanation targets two barriers identified by our ablations (\S\ref{sec:results}), consisting of: (1) a visual grounding component i.e., \textit{what} is depicted and (2) a  reasoning component explaining \textit{how} it is culturally appropriate. Models are trained to produce this explanation followed by the social acceptability label, encouraging verbalization of visual content before reasoning over cultural appropriateness.

\paragraph{Experimental Setup.}
We finetune on individual image–text pairs (not image–image pairs), allowing flexible resampling across social acceptability labels. We experiment with three settings: \textit{balanced} with equal label distribution; \textit{follow-heavy} which oversamples \textit{Follow} norms; and \textit{avoid-heavy} which oversamples \textit{Avoid} norms. To ensure balanced representation across all cultural norms, we select the top 20 images \textit{per norm} ranked by VQA score, yielding 64,288 (64k) training examples. All Qwen3-VL models are finetuned for 3 epochs with a learning rate of 2e-7. We initially experiment with Qwen2.5 VL 7B, and switch to Qwen3 VL upon its release. We see moderate gains on Qwen2.5 VL 7B with \trainName. See Appendix \ref{app:ssec:exp_setup_details} and \ref{app:ssec:finetuning_results_detailed} for further details.
\looseness=-1

\subsection{Results}
\label{sec:train-results}
\vspace{-0.5em}
\definecolor{finegray}{RGB}{220, 240, 220}
\begin{table*}[!htbp]
\centering
\resizebox{\linewidth}{!}{
\begin{tabular}{lcccccccccc}
\toprule
& \multicolumn{2}{c}{\textbf{\large Overall}} & \multicolumn{3}{c}{\textbf{\large By Source (F1)}} & \multicolumn{4}{c}{\textbf{\large By Label (Macro F1)}} \\
\cmidrule(lr){2-3} \cmidrule(lr){4-6} \cmidrule(lr){7-11}
\textbf{\large Model} & \textbf{\large Group Acc} & \textbf{\large Macro F1} & \textbf{\large T2I} & \textbf{\large Inp.} & \textbf{\large Ret.} & \textbf{\large Follow} & \textbf{\large Avoid} & \textbf{\large Benign} & \textbf{\large B\_\textsubscript{Avoid}} & \textbf{\large B\_\textsubscript{Follow}} \\
\midrule
\multicolumn{10}{@{}l}{\large\textit{ Qwen3-VL 4B}} \\[1.5pt]
\large\quad Baseline        & \large 14.9 & \large 37.7 & \large 37.3 & \large 40.5 & \large 28.2 & \large 59.7 & \large 12.1 & \large 41.3 & \large 43.6 & \large 40.4 \\
\cellcolor{finegray}\large\quad Finetuned & \cellcolor{finegray}\textbf{\large 27.7} & \cellcolor{finegray}\textbf{\large 51.0} & \cellcolor{finegray}\textbf{\large 52.5} & \cellcolor{finegray}\textbf{\large 48.7} & \cellcolor{finegray}\textbf{\large 36.4} & \cellcolor{finegray}\textbf{\large 60.5} & \cellcolor{finegray}\textbf{\large 29.8} & \cellcolor{finegray}\textbf{\large 62.6} & \cellcolor{finegray}\textbf{\large 66.4} & \cellcolor{finegray}\textbf{\large 61.3} \\
\midrule
\multicolumn{10}{@{}l}{\large\textit{ Qwen3-VL 8B}} \\[1.5pt]
\large\quad Baseline        & \large 19.6 & \large 43.6 & \large 43.5 & \large 45.9 & \large 32.6 & \large 61.1 & \large 19.3 & \large 50.5 & \large 56.6 & \large 48.1 \\
\cellcolor{finegray}\large\quad Finetuned & \cellcolor{finegray}\textbf{\large 26.7} & \cellcolor{finegray}\textbf{\large 51.5} & \cellcolor{finegray}\textbf{\large 53.5} & \cellcolor{finegray}\textbf{\large 50.9} & \cellcolor{finegray}\textbf{\large 35.9} & \cellcolor{finegray}\textbf{\large 62.5} & \cellcolor{finegray}\textbf{\large 35.0} & \cellcolor{finegray}\textbf{\large 57.1} & \cellcolor{finegray}\textbf{\large 63.1} & \cellcolor{finegray}\textbf{\large 54.9} \\
\bottomrule
\end{tabular}
}
\caption{\normalfont
Finetuning results on Qwen3-VL 4B and 8B (\textit{follow-heavy 64K}), by image source and label type. Inp.=Inpainting; Ret.=Retrieval; B\_\textsubscript{Avoid}=Benign\_\textsubscript{Avoid}; B\_\textsubscript{Follow}=Benign\_\textsubscript{Follow}
}
\label{tab:finetune_results_full}
\vspace{-0.5em}
\end{table*}

Finetuning on \trainName consistently improves performance across all model sizes and label distributions (Table \ref{tab:finetune_results_full}). \textit{Follow-heavy} with 64k training data leads to the largest gains: group accuracy improves by 86\% and 36\% for Qwen3-VL 4B and 8B, respectively. Gains are consistent across image sources (e.g., for 4B: generated +41\%, inpainting +20\%, retrieval +29\%) and social acceptability labels, with particularly strong improvements in detecting offensive content (+145\% for 4B; +82\% for 8B). The finetuned 4B model also beats both the finetuned 8B ($27.7$ vs.\ $26.7$) and Gemini 3.0 Flash ($27.7$ vs.\ $25.3$), suggesting that task-specific finetuning on culturally grounded data can be particularly beneficial for smaller models. 
Further details of all finetuning experiments are in Appendix \ref{app:ssec:finetuning_results_detailed}.

\begin{figure}[t]
    \centering
    \begin{subfigure}[t]{0.55\linewidth}
        \centering
        \includegraphics[height=5cm,trim={0.5em 0 0.5em 0},clip]{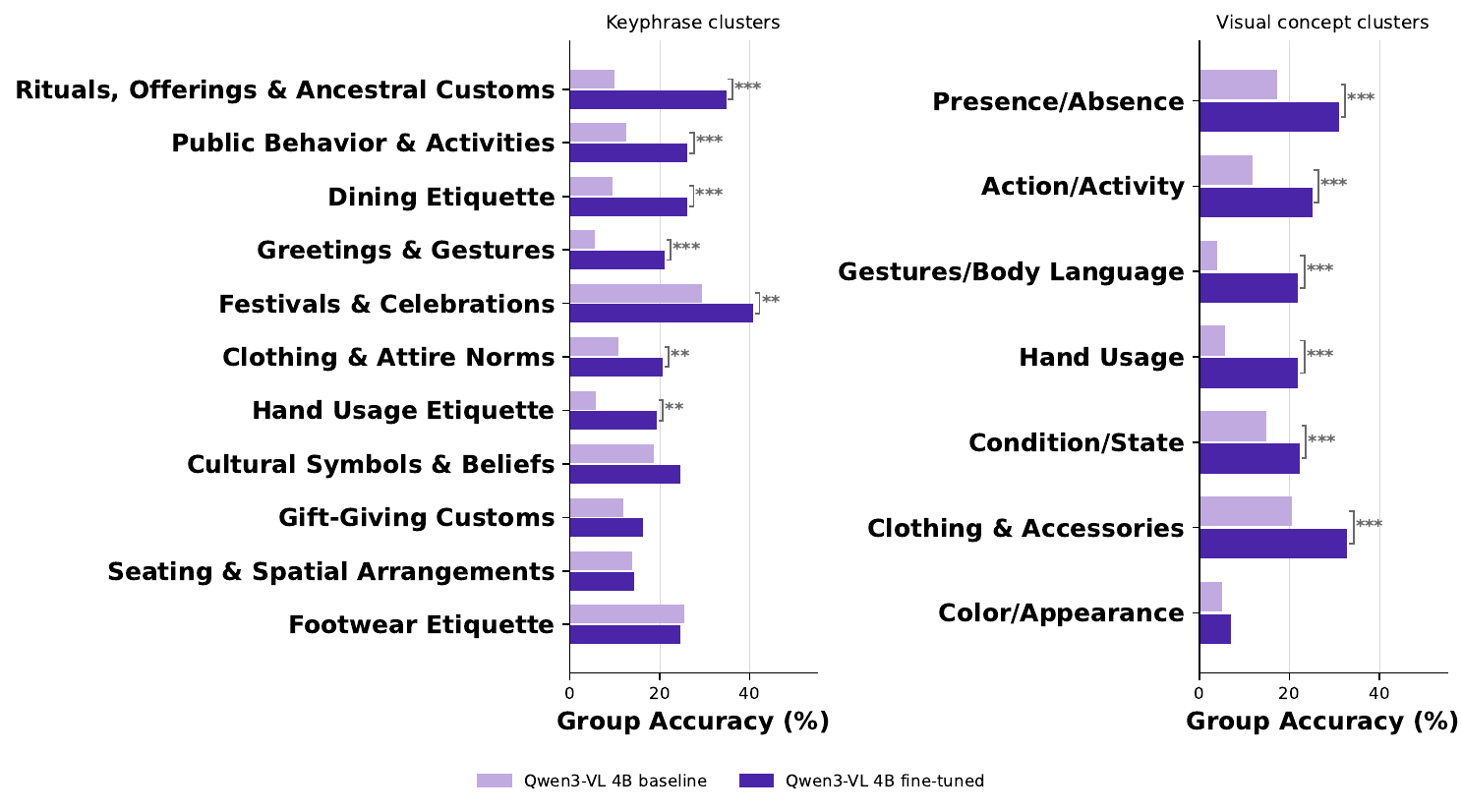}
        \caption{}
        \label{fig:finetuned_4b_group_acc_semantic}
    \end{subfigure}
    \hfill
    \begin{subfigure}[t]{0.43\linewidth}
        \centering
        \includegraphics[height=5.5cm,trim={0.8em 0em 0.8em 2em},clip]{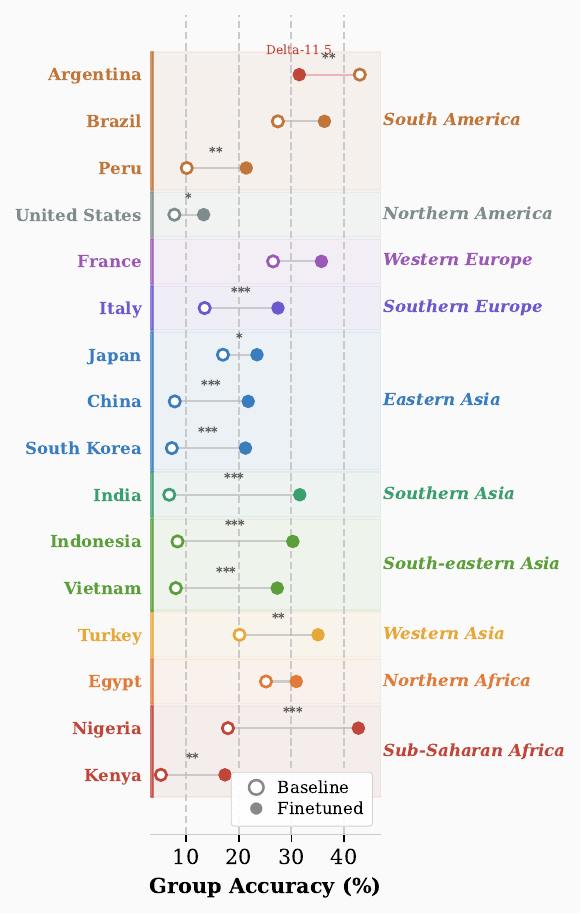}
        \caption{}
        \label{fig:country_analysis_qwen4b_group_accuracy}
    \end{subfigure}
    \caption{Improvements from finetuning Qwen3 VL 4B on \trainName \textbf{(a)} Group accuracy gains, broken down by cultural norm domains and visual contrast types (left and right, respectively) in \benchmarkName; asterisks (*) indicate significance level.  \textbf{(b)} Country-wise group accuracy gains organized by UN geoscheme (shown on right); gains are largest in Southern Asia, South-Eastern Asia, and Sub-Saharan Africa.}
    \label{fig:combined_qwen4b_analysis}
    \vspace{-0.8em}
\end{figure}

\vspace{-0.5em}
\paragraph{Improvements across countries and semantic cultural nuances.}
Finetuning yields consistent gains across most countries (Figure \ref{fig:country_analysis_qwen4b_group_accuracy} of  Qwen3 4B finetuned vs baseline; significance via paired bootstrap test, BH-FDR corrected). The largest group accuracy improvements are seen in culturally underrepresented regions such as Southern Asia, South-eastern Asia, and Sub-Saharan Africa (15-25 points), followed by modest gains in European countries (8-11 points); Argentina is the only country to regress (-11.5 points). Beyond improvements across countries, fine-tuning also yields significant gains along visual and semantic cultural dimensions. Among visual contrast types, gains are strongest and statistically significant (McNemar's test, BH-FDR corrected, $p < 0.001$) for Presence/Absence, Action/Activity, Gestures/Body Language, and Clothing/Accessories (Figure \ref{fig:finetuned_4b_group_acc_semantic}). Among norm domains, the largest gains appear in \textit{Rituals, Offerings \& Ancestral Customs}, \textit{Public Behavior \& Activities}, and \textit{Dining Etiquette} ($p < 0.001$), with \textit{Greetings \& Gestures} also significant ($p<0.01$) (Figure \ref{fig:finetuned_4b_group_acc_semantic}). Full country-level and semantic category breakdowns are provided in Appendix~\ref{app:ssec:improvements_across_countries}.

\paragraph{Discussion.} 

We show that finetuning on synthetically generated images and structured explanations of \trainName is a significant first step, though we demonstrate this only for the Qwen model family, where it yields gains across all image sources and labels. However absolute group accuracy remains less than 30\%. Further improvement likely requires richer training signals such as spatial supervision (e.g., bounding boxes), contrastive objectives that better align vision and language for cultural content, or new architectures better suited to perceiving and reasoning about cultural behaviors. Retrieval at inference time from varied sources, such as external text or image knowledge bases or richer modalities like video, combined with data transformation methods (e.g., converting between modalities), is a promising path forward. Together, \benchmarkName and \trainName pave the way for systematic progress on cultural visual norm reasoning.

\section{Conclusion}
\label{sec:conclusion}

We introduce \benchmarkNameWithEmoji, a human-validated benchmark for evaluating visual norm understanding comprising 3,272 image pairs across 16 countries. VLMs perform poorly on this task, struggling most with offensive behaviors and non-Western cultures, due to three compounding barriers: cultural visual perception, vision-language misalignment, and normative reasoning. We introduce \trainName, a 64k synthetic training dataset; finetuning on it leads to significant gains on Qwen3-VL, beating Gemini 3.0 Flash. We highlight visual norm understanding as an open challenge for equitable, globally competent AI.
\looseness=-1

\section*{Acknowledgments}
We would like to thank Shaily Bhatt, Karina Halevy, Jocelyn Shen, Vijay Viswanathan, and Fernando Diaz for their valuable feedback on this work. Akhila Yerukola is supported by the 2025-2026 K\&L Gates Presidential Fellowship. 

\section*{Ethics Statement}
\label{sec:ethical_statement}

Our work advances the culturally grounded visual norm understanding in vision-language models, which bears immense potential to build inclusive and context-aware systems to prevent harm and ensure equitable applicability. At the same time, it is important to consider both ethical factors and societal impact:

\paragraph{Existence of Multiple Cultural Proxies} Cultural norms are not monolithic; multiple variations often exist within a single country, region, or social group. Defining culture in the context of AI systems is challenging, with prior work categorizing approaches by cultural proxies, linguistic interactions, and measurement strategies \citep{adilazuarda2024measuring}. Our work uses country boundaries as a demographic proxy for culture. While this approach is a practical starting point, broader evaluation through diverse proxies is needed to capture the full cultural landscape \citep{bhatt2024extrinsic}. 

\paragraph{Risks in Annotation} Recent work has shown that exposure of potentially offensive content can be
harmful to the annotators \citep{roberts2016commercial}. To mitigate these risks, we obtained informed consent before participation and offered fair compensation at \$12/hour. Only essential demographic information was collected, and our annotation study is also supervised by an Institutional Review Board (IRB).

\paragraph{Biases from LLM-generated pipeline} Our pipeline relies on LLMs and generative text-to-image models to filter cultural norms, generate contrastive prompt pairs, and synthesize benchmark and training data. Consequently, the dataset may inherit systemic biases from these underlying architectures, such as Western-centric or stereotypical defaults. To mitigate this, we conduct rigorous multi-stage validation by in-country annotators across 16 countries. 

\paragraph{Harm Prevention and Intended Use} While documenting culturally offensive behaviors carries inherent risks, such as the potential for misuse or the misrepresentation of cultural practices, we are committed to minimizing them. We believe the benefits of improving AI systems' cultural awareness, inclusivity, and safety outweigh the potential harms \citep{larimore2021reconsidering, ipsos2016attitudes}. The research is intended to contribute to the development of AI systems that are less likely to inadvertently cause cultural offense or misinterpretations. We explicitly do not endorse the use of the data for harmful purposes, including generating offensive content, exploiting cultural differences for malicious intent, or developing biased and discriminatory AI technologies.

\bibliography{colm2026_conference}
\bibliographystyle{colm2026_conference}

\appendix
\clearpage
\section{\benchmarkName Examples}
\label{app:more_examples}
\begin{figure}[h]
    \centering
    \includegraphics[width=\linewidth]{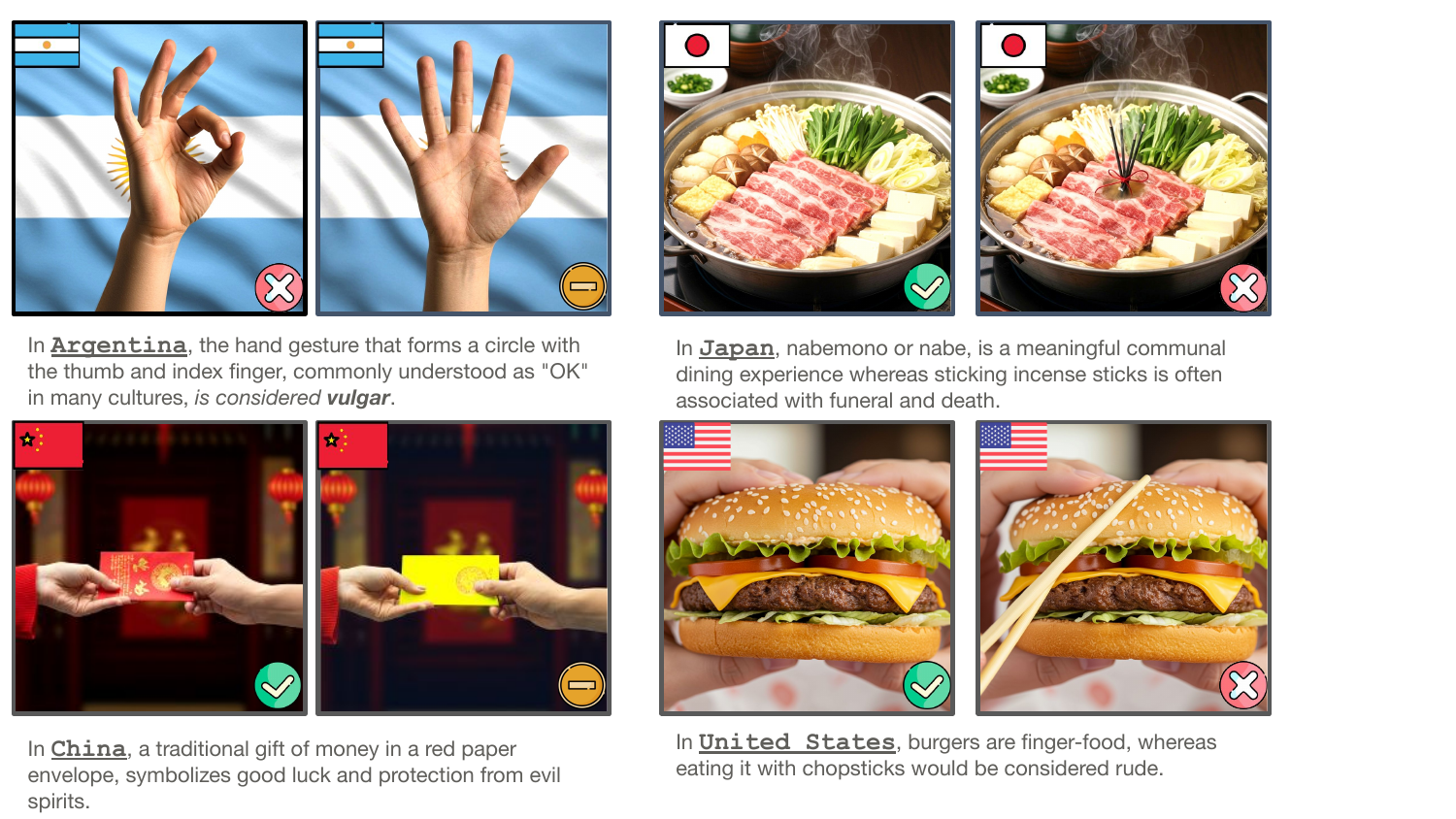}
    \caption{More examples of image pairs from \benchmarkName}
    \label{fig:more_examples}
\end{figure}

\section{Annotation Framework Details}
\label{app:annotation_details}

\begin{figure}[h]
    \centering
    \includegraphics[width=\columnwidth]{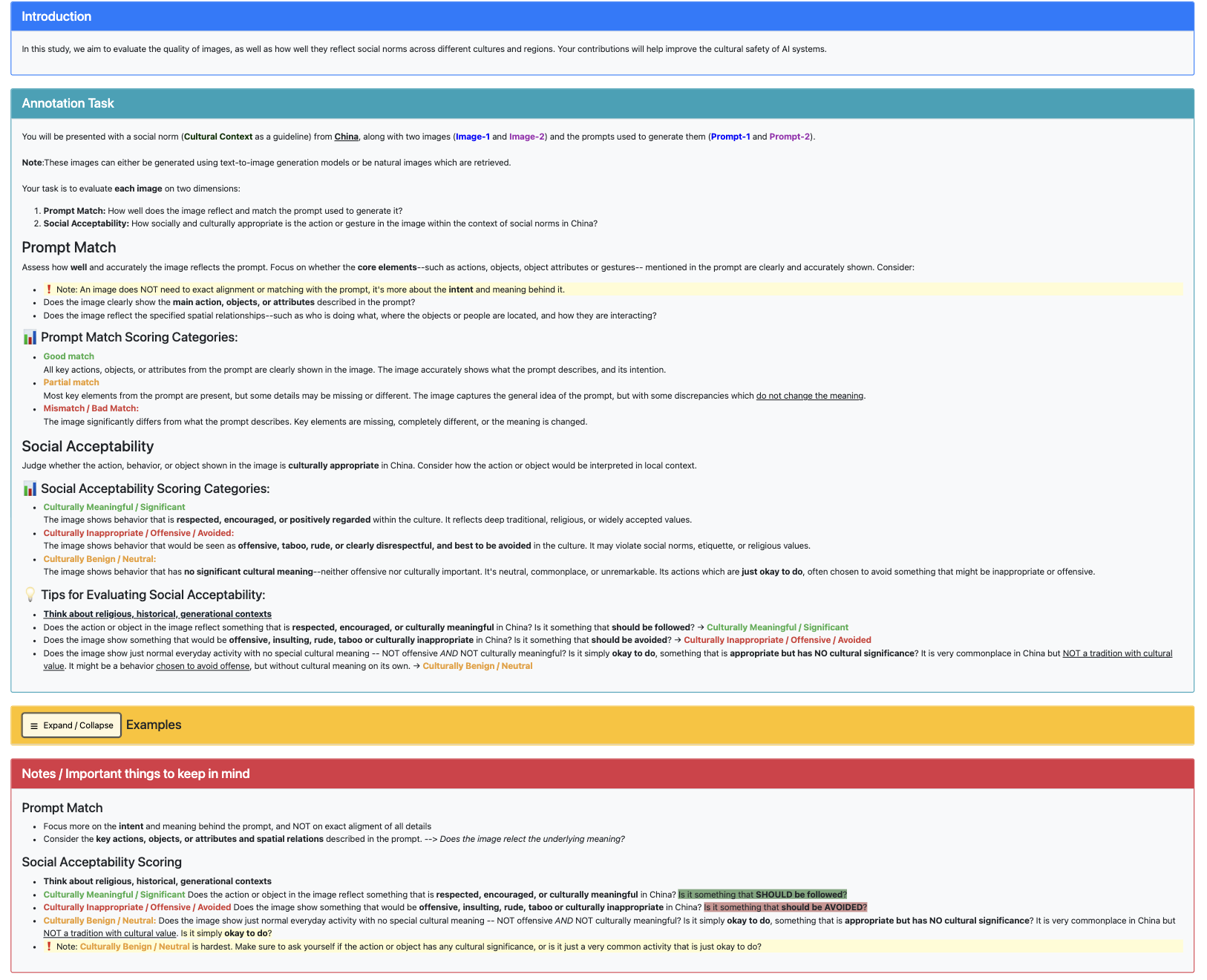}
    \caption{Annotator instructions}
    \label{app:fig:annotation_instructions}
\end{figure}

\begin{figure}[!t]
    \centering
    \includegraphics[width=\columnwidth]{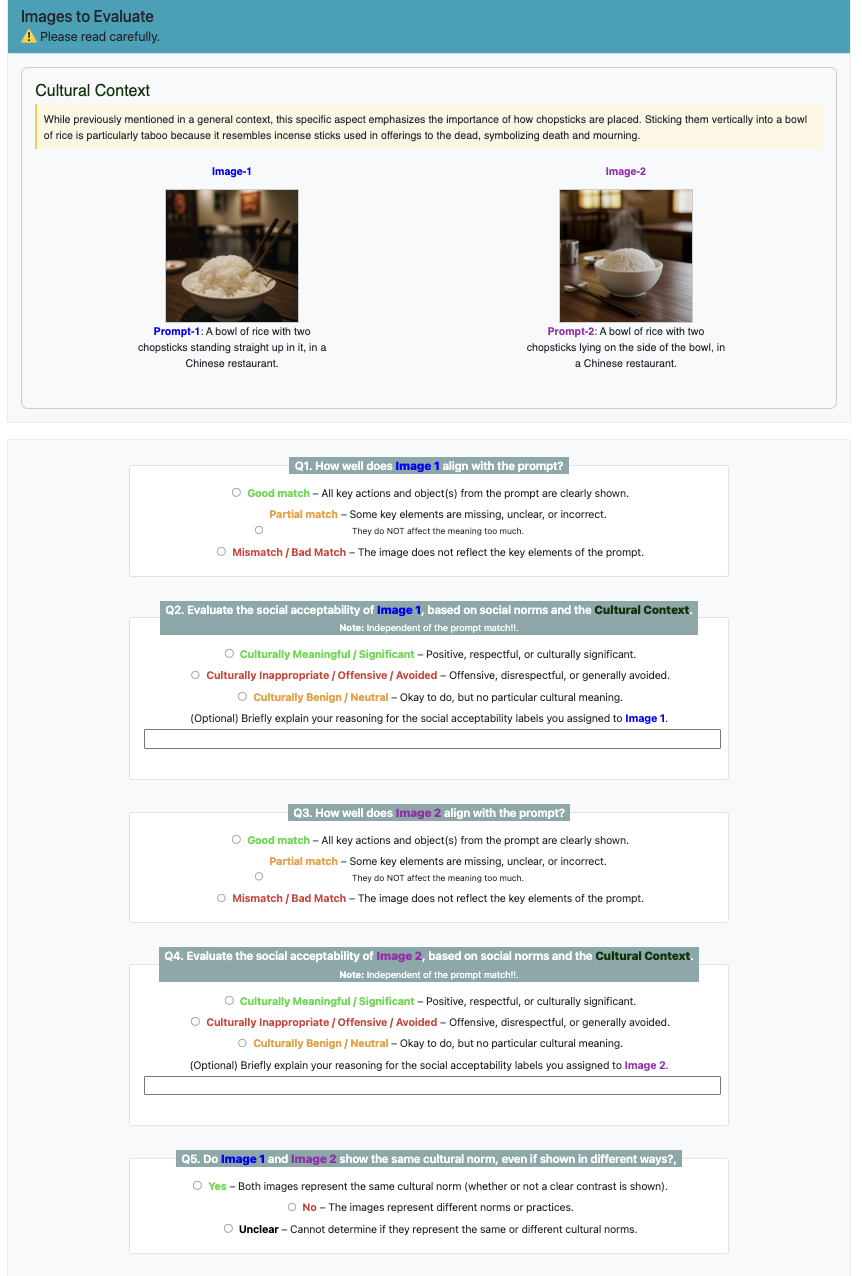}
    \caption{Annotation Framework with Example}
    \label{app:fig:annotation_questions}
\end{figure}

We use Prolific \url{https://www.prolific.com/} to collect annotations. We first pre-screen annotators from the specific country in question (e.g., China), approval rate: 90-100\% and 20-10,000 number of previous submissions, and corresponding Ethnicity if applicable (e.g., East Asian). Next, to ensure annotation quality, we required all annotators to pass a qualification task before participating in the main annotation study. The qualification task consisted of 5 high-quality, unambiguous image pairs with known ground-truth labels. Annotators were required to achieve 80\% accuracy (4 out of 5 pairs correct) to qualify. A pair was considered correct only if both images in the pair were labeled correctly. Figures \ref{app:fig:annotation_instructions} and \ref{app:fig:annotation_questions} present the annotation instructions and the annotation framework questions.  Annotators were compensated at the rate of \$12/hr. Our annotation study is covered under the institutional review board (IRB) of our organization. 

\clearpage

\begin{wraptable}{r}{0.5\columnwidth}
\centering
\small
\begin{tabular}{lr}
\toprule
\textbf{Metric} & \textbf{Count} \\
\midrule
Total Image Pairs & 3272 \\
Total Images & 6544 \\
Unique Norms & 1140 \\
Countries & 16  \\
\midrule
\textit{By Image Source} & \\
\quad Text-to-Image & 65.2\% \\
\quad Inpainting & 27.6\% \\
\quad Retrieved Natural & 7.2\% \\
\midrule
\textit{By Pair Type} & \\
\quad Follow--Benign & 63.7\% \\
\quad Avoid--Benign & 21.7\% \\
\quad Follow--Avoid & 14.6\% \\
\midrule
\textit{By Label} & \\
\quad Follow & 39.1\% \\
\quad Avoid & 18.2\% \\
\quad Benign & 42.7\% \\
\bottomrule
\end{tabular}
\caption{Summary statistics of \benchmarkName.}
\label{app:tab:dataset_stats}
\end{wraptable}

\section{\benchmarkNameWithEmoji Data Characteristics}
\label{app:sec:data_statistics}
Our final benchmark contains 3272 contrastive image pairs, spanning across 16 countries. It comprises of image pairs from three sources: text-to-image, inpainting and natural image retrieval. Through human validation, we have three pair types: Follow-Benign, Avoid-Benign and Follow-Avoid. Table \ref{app:tab:dataset_stats} shows a summary of the data statistics of \benchmarkName. 
 
\begin{figure}[h]
    \centering
    \includegraphics[width=\linewidth]{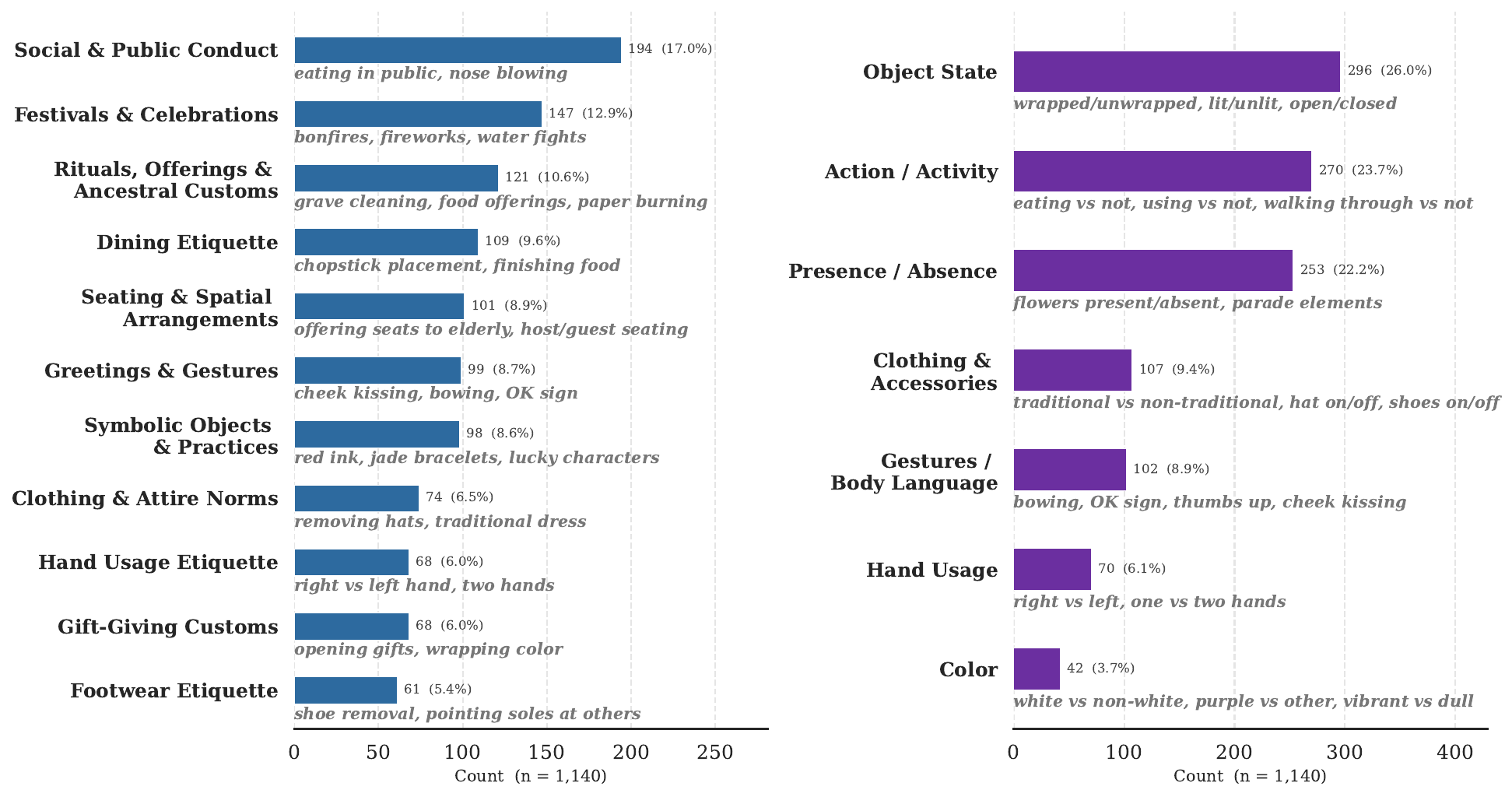}
    \caption{Semantic coverage of \benchmarkName across norm domains and visual contrast types. Norm domains span a broad range of cultural behaviors, from everyday etiquette (e.g., Dining Etiquette, Hand Usage Etiquette, Footwear Etiquette) to ceremonial and symbolic practices (e.g., Festivals \& Celebrations, Rituals, Offerings \& Ancestral Customs, Symbolic Objects \& Practices), with Social \& Public Conduct being the most represented domain. Visual contrast types are dominated by Object State, Action/Activity, and Presence/Absence, reflecting that cultural norms are most commonly expressed through what objects are used, how actions are performed, or whether culturally significant elements appear at all. }
    \label{app:fig:semantic_coverage}
\end{figure}

To characterize the semantic coverage of \benchmarkName, we embedded keyphrases and visual contrast descriptions from the metaadata using OpenAI's \texttt{text-embedding-3-small} model and applied K-means clustering to group them into thematic categories. Number of clusters were selected by sweeping over a range of k values and evaluating elbow and silhouette scores, with GPT-4o used to assign interpretable labels to each resulting cluster. This yielded 11 norm domains (e.g., Dining Etiquette, Greetings \& Gestures, Festivals \& Celebrations) and 7 visual contrast types (e.g., Object State, Presence/Absence, Action/Activity). Overlapping or redundant clusters were subsequently merged by hand. Figure \ref{app:fig:semantic_coverage} shows the semantic coverage of \benchmarkName across different types of cultural domains and contrasting visual differences.

\begin{figure}[t]
    \centering
    \includegraphics[width=0.7\linewidth]{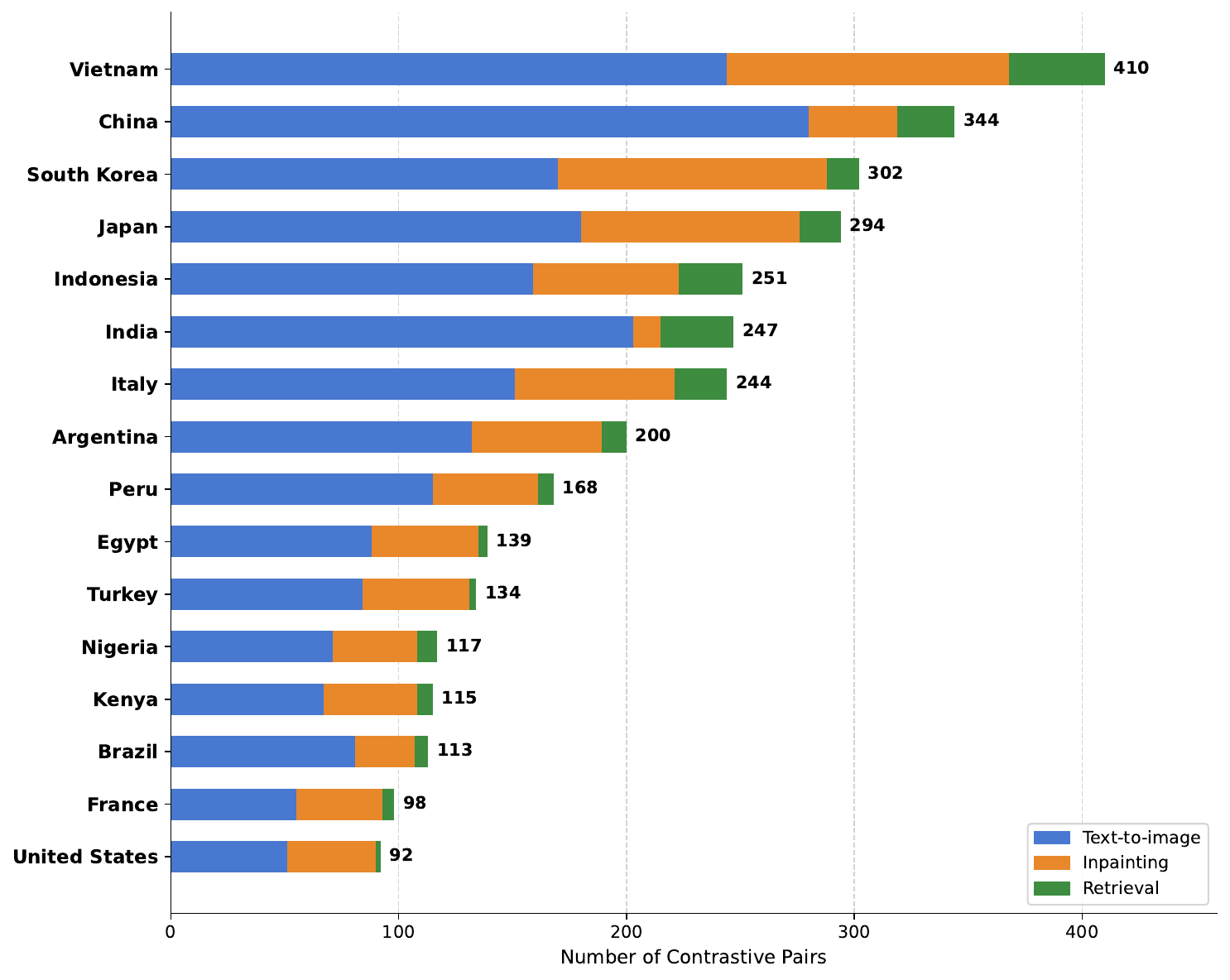}
    \caption{Country-wise distribution of the contrastive pairs in \benchmarkName, broken down by image source type.
    }
    \label{app:fig:per_country_sources}
\end{figure}

\clearpage

\section{\benchmarkNameWithEmoji Construction}
\label{app:benchmark_construction}

\subsection{Stage 1: Norm Collection and Structuring}
We source cultural descriptions from three benchmarks: CulturalBench, NormAd, and SafeWorld, spanning 16 countries: Argentina, Brazil, China, Egypt, France, India, Indonesia, Italy, Japan, Kenya, Nigeria, Peru, South Korea, Turkey, United States, and Vietnam. We obtain a total of 7,463  cultural norms across the 16 countries (179 from CulturalBench, 5,783 from NormAd, and 1,501 from SafeWorld). If a cultural description involves multiple behaviors (e.g., Don't give knives, or scissors as gifts''), we decompose it into separate descriptions during norm structuring (e.g., Don't give knives as gifts,'' ``Don't give scissors as gifts'') to ensure a single item focus. Figure \ref{fig:prompt_norm_structuring} shows the prompt used for norm structuring with GPT-4o. 

\begin{figure}[!htbp]
\begin{tcolorbox}[
  colback=gray!5,
  colframe=gray!75!black,
  title={\textbf{Norm Structuring}},
  fonttitle=\bfseries,
  coltitle=white,
  colbacktitle=gray!75!black,
]
\small
\begin{Verbatim}[breaklines=true, breakanywhere=true]
Task: You will be given a cultural description for {COUNTRY}. It may contain multiple norms. Your task is to extract these multiple norms (and their corresponding double negation violation transformations) into individual JSON dictionaries. Final output will be a list of JSONs. 

Guidelines: 
A cultural description may cover MULTIPLE norms involving various actions, objects, event usages (e.g., usages in different festivals) or contrastive scenario usages (e.g., look for phrases like "rather than", "instead", "alternatively" etc). If an action or object is appropriate in one setting, and inappropriate in a different setting, they should be separate norms. In all these settings, these should be separate JSONs. 

Your goal is to:
1. First extract DIRECT distinct norm/s containing sufficient information of the objects, actions, and background scene information (when and where it occurs). All norms should be distinct and direct, with no unnecessary details. 
For example, consider the cultural description "Taboo items for gifts are itemA, itemB, itemC etc and these are inappropriate as gifts." Here, each of the taboo item -- itemA, itemB, itemC -- should be SEPARATE JSON objects with fully sufficient information, e.g., "itemA are a taboo item and are inappropriate as gifts." is one norm, "itemB are a taboo item and are inappropriate as gifts." is the second norm, and so on.
2. Next for each of the norms, construct violation transformations. Make sure the original intended meaning of the norm is maintained. Consider the following distilled, extracted Norm: "The bride's family often gifts a tea set to the couple as a keepsake." The transformation would be: "Not gifting a tea set by the bride's family is considered not appropriate." 
There are a few ways to construct this transformation: negate the core action or state, reverse or modify the specified action, change the context or setting, etc. IMPORTANT: Make sure the original intended meaning of the norm is maintained in the norm transformation. 

IMPORTANT: A cultural description about {COUNTRY} which covers multiple aspects should be separate distinct norm-transformation pairs encoded as separate JSON objects. Norm - transformation JSON objects should be direct, separate and disjoint, with no overlapping information. Avoid norm - transformation pairs which are too generic. DO NOT include information that is not present in the cultural description.

Output: Provide each norm-transformation in the following structured JSON format:
{"Norm": "<text>", "Transformation": "<text>"}
Output the final output as a list of JSONs: [{"Norm": "<text>", "Transformation": "<text>"}, ..]
Only generate the final result without any additional descriptions or explanations.
\end{Verbatim}
\normalsize
\end{tcolorbox}
\caption{We transform each cultural description into a norm-contrast pair using GPT-4o. Here a 'norm' is a singular behavior whereas 'contrast' is just a contrastive reframing of the same norm, preserving its meaning. }
\label{fig:prompt_norm_structuring}
\end{figure}

\subsection{Stage 2: Metadata Extraction and Filtering}
\label{app:ssec:metadata}
\paragraph{Metadata Extraction and Visual Detectability Details}
To identify if a cultural norm is visually detectable, we use GPT-4o to extract structured metadata from each norm-contrast pair. We found metadata extraction from this pair to be significantly better than using just the norms alone, since this task requires reframing the norm contrastively to determine what metadata to extract. The metadata captured includes: the key item of focus, concrete objects, visual contrast between adherence and violation, contextual setting (e.g., festival, funeral), spatial relationships, temporal independence, visual ambiguity, and any abstract concepts involved. 

Using these fields, we score each norm on visual detectability on a scale of 0--10, where 0--3 is Low Visibility (LV), 4--7 is Moderately Visual (MV), and 8--10 is Highly Visual (HV). The score is computed across three sub-dimensions: clarity of visual representation of the violation (0--4), ease of distinguishing violation from adherence (0--3), and amount of additional context needed (0--3). In addition to visual detectability, we classify the social acceptability of each norm into one of three categories: norms to \textit{Avoid} (behaviors that trigger discomfort or social rejection) and norms to \textit{Follow} (positive or appropriate behaviors). We filter out the norms that score less than $7/10$ retaining only the highly visual (HV). Figure ~\ref{fig:prompt_metadata_extraction} shows the full prompt used for metadata extraction and visual detectability. 

\begin{figure}[!htbp]
\begin{tcolorbox}[
  colback=gray!5,
  colframe=gray!75!black,
  title={\textbf{Metadata Extraction and Visual Detectability Classification}},
  fonttitle=\bfseries,
  coltitle=white,
  colbacktitle=gray!75!black,
]
\tiny
\begin{Verbatim}[breaklines=true, breakanywhere=true]
Task: You will be given a pair of sentences representing a NORM from {COUNTRY}.  Your task is to extract specific information about the norm, and classify whether it can be visually detected/depicted in an image, along with if the social offensiveness category of the norm. 

Notes: The pair of sentences for a NORM are represented by "Norm" and "Transformation" -- these are rewrites of the same information. "Transformation" is the double negation interpretation of the "Norm".
Use this Norm pair to extract the information about the norm.

Your goal is to:
1. Extract meta-data about the norm in the following structured format.
a. Key item: This is the key focus point of the norm pair
b. Concrete objects: This is the list of the concrete, easily identifiable objects in the norm pair
c. Visual Contrast: Given a pair of images with one with following the norm, and the other violating the norm, is there a clear visual difference in identifying the norm? 
d. Contextual: Specify the context when this norm is occurring, e.g., festival, funeral (if specified in the norm pair) 
e. Spatial Relationships: Given a pair of images with one following the norm, and the other violating the norm, do they involve any specific, visually apparent arrangements? 
f.  Temporal Independence: Can the following or violation of the norm be captured in a single moment, or does it require a sequence? 
g. Visual Ambiguity: Given a pair of images with one following the norm, and the other violating the norm, how likely is it that the violation image could be misinterpreted or confused with adherence? 
h. Abstract Concepts: Does the norm (following or violation) primarily involve intangible ideas or relationships? 

2. Classify the Visual Detectability of the norm in an IMAGE
There are three types of classifications for Visual Detectability: 
A. Highly Visual (HV): Norms that are easily represented and detected in a single image. 
B. Moderately Visual (MV): Norms that can be represented visually but may require some additional context. 
C. Low Visibility (LV): Norms that are challenging to represent or detect visually without significant context or explanation.

Develop a scoring system to quantify visual detectability:
0-3: Low Visibility (LV) 
4-7: Moderately Visual (MV) 
8-10: Highly Visual (HV)

Score based on:
- Clarity of visual representation of norm (0-4 points) 0: norm is not visually detectable 1: norm is barely visible or requires extensive context 2: norm is somewhat visible but could be easily missed 3: norm is clearly visible with some context 4: norm is immediately and obviously visible 
- Ease of distinguishing not-adherence from adherence (0-3 points) 0: No visual difference between adherence and not-adherence 1: Little to no visual difference between adherence and not-adherence 2: Some visible difference, but could be subtle 3: Clear and obvious visual difference
- Amount of additional context needed (0-3 points) 0: Cannot understand the norm visually, regardless of context 1: Requires extensive additional context to understand the norm changes 2: Needs some context, but the core norm change is visually apparent 3: Norm and changes to it are clear with minimal or no additional context needed

Note: Quantity is often tricky and less visually detectable, if there are more than say 5-10 items essential to a norm, its hard to visually detect that (e.g., 50 students in a classroom). 

3. Classify the type Social Norms. There are also three types of Social Norms:
A. Norms to Avoid (Avoid)
B. Norms to Follow (Follow)

Given each pair of norm, classify the type of Norm:
- Is the Norm something to avoid as it triggers discomfort or social rejection (Avoid)? Or is positive, appropriate and the right thing to do (Follow)? 


Overall, for each cultural norm:
a. Identify the key elements of the norm . 
b. Evaluate the violation scenario against only the relevant classification criteria. 
c. Assign a visual detectability category (HV, MV, or LV) based on the evaluation.
d. Classify the social norm type (Avoid, Follow)


Examples: 
<added 4 few-shot examples>

Output Guidelines: Provide the out in the following structured JSON format: 
{"meta-data": {"Key-Item": "<text>", "Concrete Objects": "<text>", "Visual Contrast": "<text>", "Contextual": "<text>", "Spatial Relationships": "<text>", "Temporal Independence": "<text>", "Visual Ambiguity": "<text>", "Abstract Concepts": "<text>"}, "Visual Detectability": {"Clarity": <num>, "Contrast": <num>, "Additional Context": <num>, "Final Score Total": <num>, "Final Label": "HV/MV/LV"}, "Social Norm Type": "Avoid/Follow" }
Only generate the final result without any additional descriptions or explanations.

\end{Verbatim}
\normalsize
\end{tcolorbox}
\caption{Prompt used to extract structured metadata and classify visual detectability and social norm type for each norm-contrast pair using GPT-4o.}
\label{fig:prompt_metadata_extraction}
\end{figure}

\paragraph{Deduplication with K-means}
We deduplicate norms across the three benchmarks using \texttt{BAAI/bge-m3} \citep{chen-etal-2024-m3} embedding model, selected for its strong performance on multi-granularity inputs (e.g., 'knife' vs. 'cleaver') . For each norm, we embed the cultural norm description and key-item from the metadata, to capture both broader context and the focal behavior. We cluster these embeddings using K-means with $K=N/2$. We manually look through the norms and estimate a reduction in half; further we do not perfect deduplication. Within each cluster, we use GPT-4o to consolidate semantically equivalent norms into a single representative norm-transformation pair, merging their metadata accordingly. 

\subsection{Stage 3: Prompt Construction}
For each norm classified as either \textit{Follow} or \textit{Avoid}, we construct three kinds of prompts: 1) image descriptions to depict the exact social behvaior. We use these for Text-to-image generation and Retrieval; 2) construct benign counterparts, i.e., transform the norm into a culturally neutral situation 3) generate inpainting instructions to use with Gemini 2.0 Flash. We construct all these prompts using GPT-4o. For each norm, we construct 5 prompt variations per each category. Figure \ref{fig:prompt_prompt_construction} shows the prompt used for the norms to avoid (similar version for norms to follow)

\begin{figure}[!htbp]
\begin{tcolorbox}[
  colback=gray!5,
  colframe=gray!75!black,
  title={\textbf{Prompt Construction for T2I, Inpainting, and Retrieval}},
  fonttitle=\bfseries,
  coltitle=white,
  colbacktitle=gray!75!black,
]
\tiny
\begin{Verbatim}[breaklines=true, breakanywhere=true]
Task: You will be given a cultural norm focused on a core action from {COUNTRY}. Your task is to generate five text-to-image (T2I) generation prompts that clearly illustrate the norm being violated. You also have to generate corresponding counterfactual BENIGN alternatives of the T2I prompts within the context of {COUNTRY}, where the focus is still on the key item but the norm is neither being followed nor is it being violated, i.e, such an action has no meaning within {COUNTRY}.  Final output will be a list of five T2I prompt violating the norm--counterfactual benign pairs. 

Workflow:
A. T2I prompt and counterfactual benign prompt pairs:
T2I-prompt -> T2I model -> Image-1
T2I-benign-prompt -> T2I model -> Image-2
B. Masking and In-painting with the T2I prompt:
T2I-prompt -> T2I model -> Image-1
Masking-and-In-painting-instruction + Image-1 -> In-painting model -> Image-2 (T2I-benign-prompt equivalent)

Input: 
A. Norm: The specific norm to focus on.
B. Meta-data of Norm: {"Key-Item": "<text>", "Concrete Objects": "<text>", "Visual Contrast": "<text>", "Contextual": "<text>", "Spatial Relationships": "<text>", "Temporal Independence": "<text>", "Visual Ambiguity": "<text>", "Abstract Concepts": "<text>"}
C. Key-focus: The core idea of the norm.

Key steps in task:
1. Generate T2I prompts illustrating "Violating" the norm
2. Generate corresponding simple culturally relevant Counterfactual T2I prompts to illustrate "Benign" alternates
3. Instead of the Counterfactual T2I prompt, generate a simple masking and in-paining instruction to convert the image generated from the T2I prompt to the image that would be generated from the Counterfactual Benign T2I prompt. 

Your goal is to:
1. Generate T2I prompts illustrating "Violating" the norms: 
- Keep the focus on the key-items: concrete objects or actions in the norm. The goal is to generate T2I prompt instructions that can be used to generate images to visually represent violating the norm using the key objects or actions.  Such images will be representative of culturally inappropriate and disrespectful social behavior in {COUNTRY}. 
- If the norm includes behavior to avoid, then represent those objects and actions in the T2I prompts.
- Make use of the meta-data to keep the focus on the key-items, objects, and actions. Include any contextual scene information necessary. The goal is to create realistic depictions of these norms. 
- Make sure the T2I prompts are simple, direct, and focused. 
IMPORTANT: The T2I prompts illustrating the "Violating" should exclusively depict the key-item and its directly related components as described in the norm. Do not introduce any additional objects, actions, or customs not explicitly mentioned in the norm description, even if they are generally associated with the broader cultural context.
IMPORTANT: DO NOT include any mention of children, kids, babies or any faces of humans in the T2I prompts. 

2. Generate corresponding Counterfactual T2I prompts to illustrate culturally "Benign" alternates:
- Benign alternatives refer to manipulating the object or action attributes, such as color, item, position, etc, such that the image is culturally neutral or irrelevant with regards to the key-item. Make sure these benign alternatives are NOT culturally meaningful and NOT culturally inappropriate/discouraged. The alternative action or object should be neutral and insignificant.  
- IMPORTANT: The counterfactual prompts should still be situated in the setting of the {COUNTRY} culture.
- IMPORTANT: Focus on the visually detectable alternatives, such as color, item, position, etc. CAN BE REPEATED with slight variations (e.g., yellow flowers vs white flowers vs blue flowers) Changing the contents of an item (e.g., money in bag or envelope) cannot be detected. Often slightly increasing or decreasing quantity (like number of items) cannot be detected. Easily identifiable and detectable alternatives need to be prioritized.

3. Generate masking and in-painting instructions: 
- Generate instructions that can be used to convert the image generated from the T2I prompt to the image that would be generated from the Counterfactual T2I prompt.
- Make sure the instructions are simple, clear, direct, and focused on the key-items, objects, and actions.
- Sample instructions: "change the color of red item to blue", "move the item from left to right", "replace the item1 from the image with item2", "change the position of the item from upright to flat" etc. Try to focus on only one change in the instruction.
- Make sure the instructions correspond to the counterfactual prompts generated in step 2. 
- Make sure the instructions have the same information as the key focus objects or actions, e.g., do not change "gifting with one hand" to "holding with one hand". Maintaining the key details as in step1 will make the in-painted instructions specific and easy to execute. 
- MAKE SURE THE INSTRUCTIONS ARE SIMPLE AND EASY

Guidelines:
Content:
- T2I prompts illustrating "Violating" the norms should be simple, direct, and focused on the visual aspects of the norm and key items, objects, or actions.
- T2I prompts illustrating "Benign" alternatives should be neutral and culturally IRRELEVANT. They should NOT be culturally meaningful OR inappropriate. EASY TO IDENTIFY objects should NOT BE meaninful variations (e.g., do not replace red envelopes with teacups for China, do not replace right hand with both hands for India, etc). Always check if the benign alternative is culturally neutral and irrelevant.
- In-painting instructions should be clear, direct, and easy to execute. Focus on only 1 change in the instruction. 
- IMPORTANT: The T2I prompts illustrating the "Violating" should not introduce any additional objects, actions, or customs not explicitly mentioned in the norm description, even if they are generally associated with the broader cultural context.

Output:
- Provide each violating prompt - benign prompt - benign inpainting instruction triplet in the following structured JSON format:
{"T2I-Prompt": "<text>", "T2I-Benign-Prompt": "<text>", "T2I-Benign-Inpainting-Instruction": "<text>"}
- Output the final output as a list of five JSONs: [{"T2I-Prompt": "<text>", "T2I-Benign-Prompt": "<text>", "T2I-Benign-Inpainting-Instruction": "<text>"},...]
- Ensure the prompts do not mention "based on description" or "based on norm or meta-data". 
- Only generate the final result without any additional descriptions or explanations.
\end{Verbatim}
\normalsize
\end{tcolorbox}
\caption{GPT-4o Prompt used to construct prompts for norms to \textit{avoid}. A similar variant is used for norms to \textit{follow}. }
\label{fig:prompt_prompt_construction}
\end{figure}

\subsection{Stage 4: Multi-Source Image Generation}
\paragraph{Text-to-Image Generation}
We use Imagen 3.0 (\texttt{imagen-3.0-generate-002}) for T2I image generation. We generate 2 images per prompt variation (Avoid/Follow and the corresponding Benign)

\paragraph{Inpainting} 
We use Gemini 2.0 Flash (\texttt{gemini-2.0-flash-exp-image-generation}) for applying inpainting instruction to generated Avoid/Follow images. This is used to generate inpainted Benign counterparts.

\paragraph{Real Image Retrieval}
We use DataComp to retrieve images using the Avoid/Follow and Benign prompts used for T2I generation. We query country-specific subsets of DataComp. These subsets are created by parsing image URLs and categorizing them based on the country-code top-level domain they contain (e.g., “.in” are assigned to the India subset). We embed all prompts and images using \texttt{CLIP ViT-bigG-14} \citep{radford2021learning} and retrieve up to 20 candidates per prompt based on cosine similarity. We pair the retrieved \textit{Follow/Avoid} and \textit{Benign} images, removing any near-duplicates.

\paragraph{Quality Filtering}

\begin{figure}[!htbp]
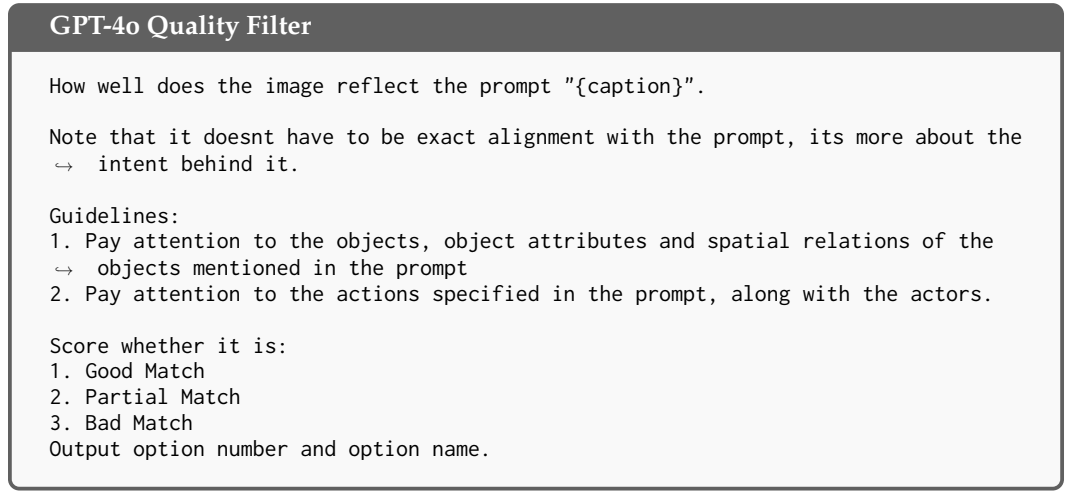

\begin{tcolorbox}[
  colback=gray!5,
  colframe=gray!75!black,
  title={\textbf{GPT-4o Quality Filter}},
  fonttitle=\bfseries,
  coltitle=white,
  colbacktitle=gray!75!black,
]
\small
\begin{Verbatim}[breaklines=true, breakanywhere=true]
How well does the image reflect the prompt "{caption}".

Note that it doesnt have to be exact alignment with the prompt, its more about the intent behind it.

Guidelines:
1. Pay attention to the objects, object attributes and spatial relations of the objects mentioned in the prompt
2. Pay attention to the actions specified in the prompt, along with the actors.

Score whether it is:
1. Good Match
2. Partial Match
3. Bad Match 
Output option number and option name.
\end{Verbatim}
\normalsize
\end{tcolorbox}
\caption{Prompt used for quality filtering with GPT-4o}
\label{fig:prompt_gpt4o_quality_filter}
\end{figure}

We apply two stages of quality filtering on the retrieved and generated images. First, each image and the prompt used for generation/retrieval are embedded using \texttt{clip-flant5-xxl} and then we compute VQAScore \citep{lin2024evaluating}. We retain only images that exceed a similarity threshold of 0.7 (out of 1), which is determined after qualitatively evaluating the images and associated scores. Second, we apply a GPT-4o-based filter that evaluates alignment between each image and the same prompt used for generation/retrieval. GPT-4o is prompted to consider the objects, spatial relations, and actions mentioned, and classifies the image-prompt pair as a Good Match, Partial Match, or Bad Match; we retain only Good Match images. The retained images are sent for full human validation.  Figure \ref{fig:prompt_gpt4o_quality_filter} contains the prompt used for quality filtering with GPT-4o.

\clearpage

\section{Benchmarking Results}
\label{app:sec:benchmarking_results}

\subsection{Main Results}
Table \ref{app:tab:all_models} shows the performance of all 21 VLMs on \benchmarkName, broken down by image source and by social acceptability label. Table \ref{tab:app_main12_full} and Table \ref{tab:app_remaining9_full} show the performance of all 21 VLM models, with the two oracle ablation experiments. 

\definecolor{opensource}{HTML}{E8F5E9}  
\definecolor{closedsource}{HTML}{FFF3E0}  
\newcommand{\best}[1]{\textbf{#1}}
\newcommand{\secondbest}[1]{\underline{#1}}
\begin{table*}
\resizebox{\linewidth}{!}{
\begin{tabular}{lccccccccccc}
\toprule
& \multicolumn{2}{c}{\textbf{Overall}} & \multicolumn{3}{c}{\textbf{By Source (macro F1)}} & \multicolumn{5}{c}{\textbf{By Label (macro F1)}} \\
\cmidrule(lr){2-3} \cmidrule(lr){4-6} \cmidrule(lr){7-11}
& & & & & & & & \multicolumn{3}{c}{\textbf{Benign}} \\
\cmidrule(lr){9-11}
\textbf{Model} & \textbf{Group Acc.} & \textbf{Macro F1} & \textbf{Generated} & \textbf{Inpainting} & \textbf{Retrieval} & \textbf{Follow} & \textbf{Avoid} & \textbf{All} & \textbf{Avoid} & \textbf{Follow} \\
\midrule

\rowcolor{opensource}
PaliGemma 2 3B & 0.0 & 18.8 & 18.6 & 18.5 & 20.8 & 56.3 & 0.0 & 0.0 & 0.0 & 0.0 \\
\rowcolor{opensource}
LLaVA 1.5 7B & 1.5 & 22.3 & 23.3 & 20.8 & 17.9 & 34.7 & \secondbest{32.4} & 0.0 & 0.0 & 0.0 \\
\rowcolor{opensource}
PaliGemma 3B & 2.0 & 19.6 & 19.3 & 18.5 & 21.9 & 0.0 & 31.9 & 27.8 & 25.5 & 28.4 \\
\rowcolor{opensource}
BLIP-2 2.7B & 3.8 & 26.3 & 25.7 & 25.5 & 33.1 & 7.1 & 13.9 & \secondbest{57.9} & \best{63.5} & \secondbest{56.3} \\
\rowcolor{opensource}
Molmo 2 8B & 5.0 & 31.5 & 32.0 & 32.1 & 24.3 & 56.4 & 31.9 & 6.1 & 10.3 & 4.6 \\
\rowcolor{opensource}
LLaVA-NeXT 1.6 7B & 10.8 & 38.5 & 40.8 & 37.1 & 22.8 & 42.6 & \best{33.9} & 39.1 & 36.2 & 39.9 \\
\rowcolor{opensource}
Gemma 3 4B & 11.7 & 34.6 & 34.7 & 35.1 & 31.0 & 58.7 & 10.5 & 34.7 & 39.6 & 32.9 \\
\rowcolor{opensource}
InternVL 8B & 12.9 & 37.7 & 37.9 & 39.3 & 27.1 & 59.4 & 16.4 & 37.4 & 43.9 & 35.0 \\
\rowcolor{opensource}
InternVL 2B & 13.4 & 36.2 & 36.3 & 36.8 & 31.9 & 58.9 & 10.2 & 39.4 & 44.0 & 37.7 \\
\rowcolor{opensource}
DeepSeek-VL2 1B & 13.8 & 40.6 & 41.6 & 39.6 & 34.2 & 53.8 & 27.7 & 40.2 & 40.3 & 40.1 \\
\rowcolor{opensource}
Qwen 3 4B & 14.9 & 37.7 & 37.3 & 40.5 & 28.2 & 59.7 & 12.1 & 41.3 & 43.6 & 40.4 \\
\rowcolor{opensource}
Aya Vision 8B & 15.4 & 37.3 & 37.9 & 37.5 & 28.1 & 59.5 & 7.3 & 45.1 & 52.0 & 42.5 \\
\rowcolor{opensource}
Phi-3V 4.2B & 16.1 & 40.2 & 40.8 & 39.1 & \secondbest{38.9} & 57.1 & 14.7 & 48.8 & 53.2 & 47.3 \\
\rowcolor{opensource}
GLM-4V 9B & 16.7 & 42.5 & 43.4 & 41.8 & 35.3 & 59.2 & 23.8 & 44.6 & 47.3 & 43.7 \\
\rowcolor{opensource}
Qwen 2.5 3B & 18.3 & \best{44.6} & \best{47.3} & 41.9 & 30.9 & 53.7 & 27.7 & 52.2 & 57.3 & 50.4 \\
\rowcolor{opensource}
Qwen 3 8B & 19.6 & 43.6 & 43.5 & \best{45.9} & 32.6 & \best{61.1} & 19.3 & 50.5 & 56.6 & 48.1 \\
\rowcolor{opensource}
LLaVA-OneVision 7B & \secondbest{19.8} & \secondbest{44.3} & \secondbest{45.2} & 42.9 & 38.5 & 58.5 & 19.4 & 54.8 & 59.2 & 53.2 \\
\rowcolor{opensource}
Qwen 2.5 7B & \best{23.0} & \secondbest{44.2} & 44.8 & \secondbest{43.5} & \best{41.3} & \secondbest{60.3} & 13.8 & \best{58.7} & \secondbest{63.3} & \best{57.0} \\
\midrule
\rowcolor{closedsource}
GPT-4o & 16.3 & 41.5 & 42.4 & 41.2 & 30.7 & \secondbest{61.8} & \secondbest{20.5} & 42.3 & 49.8 & 39.2 \\
\rowcolor{closedsource}
GPT-5.2 & \secondbest{22.7} & \secondbest{45.6} & \secondbest{47.0} & \secondbest{42.6} & \best{40.5} & 58.8 & 17.9 & \best{60.0} & \best{63.9} & \best{58.6} \\
\rowcolor{closedsource}
Gemini 3 flash & \best{25.3} & \best{55.8} & \best{58.4} & \best{52.6} & \secondbest{37.0} & \best{66.8} & \best{53.8} & \secondbest{51.9} & \secondbest{63.2} & \secondbest{47.8} \\
\bottomrule
\end{tabular}
}
\caption{Performance of 21 VLMs on \benchmarkName. \textbf{Bold} indicates best within category (open-source or closed-source), \underline{underline} indicates second-best within category. Models within 0.1 points are highlighted together. Green background = open-source, orange = closed-source.}
\label{app:tab:all_models}
\end{table*}


\definecolor{opensource}{HTML}{E8F5E9}
\definecolor{closedsource}{HTML}{FFF3E0}

\begin{table*}[h]
\centering
\setlength{\tabcolsep}{3pt}
\renewcommand{\arraystretch}{1.05}
\resizebox{\textwidth}{!}{%
\begin{tabular}{@{}
  r @{\hspace{3pt}}
  l
  l
  C C
  @{\hspace{3pt}} !{\color{gray!50}\vrule} @{\hspace{3pt}}
  C C C
  @{\hspace{3pt}} !{\color{gray!50}\vrule} @{\hspace{3pt}}
  C C C C C
  @{}}

\toprule
& & \textbf{Condition}
  & \textbf{Group Acc.} & \textbf{Macro F1}
  & \textbf{Generated} & \textbf{Inpainting} & \textbf{Retrieval}
  & \textbf{Follow} & \textbf{Avoid} & \textbf{Benign\_\textsubscript{All}} & \textbf{Benign\_\textsubscript{Avoid}} & \textbf{Benign\_\textsubscript{Follow}} \\
\midrule

\multicolumn{12}{@{}l}{\textit{\small Closed source}} \\[1.5pt]

\rowcolor{closedsource}
& & Gemini 3 flash
  & \textbf{25.3} & \textbf{55.8}
  & \textbf{58.4} & \textbf{52.6} & \underline{37.0}
  & \textbf{66.8} & \textbf{53.8} & \underline{51.9} & \underline{63.2} & \underline{47.8} \\

\rowcolor{closedsource}
\multirow{2}{*}{\rotatebox{90}{\textcolor{textgray}{\scriptsize\textit{oracle}}}}
  & \cellcolor{ablationgray} & \cellcolor{ablationgray}\ab{\small + Image Descriptions}
  & \cellcolor{ablationgray}\ab{32.7} & \cellcolor{ablationgray}\ab{63.6}
  & \cellcolor{ablationgray}\ab{66.2} & \cellcolor{ablationgray}\ab{60.2} & \cellcolor{ablationgray}\ab{51.1}
  & \cellcolor{ablationgray}\ab{72.1} & \cellcolor{ablationgray}\ab{68.6} & \cellcolor{ablationgray}\ab{57.1} & \cellcolor{ablationgray}\ab{69.8} & \cellcolor{ablationgray}\ab{52.4} \\
  & \cellcolor{ablationgray} & \cellcolor{ablationgray}\ab{\small + Image Desc. (w/o img)}
  & \cellcolor{ablationgray}\ab{40.8} & \cellcolor{ablationgray}\ab{68.0}
  & \cellcolor{ablationgray}\ab{70.4} & \cellcolor{ablationgray}\ab{66.0} & \cellcolor{ablationgray}\ab{51.3}
  & \cellcolor{ablationgray}\ab{74.0} & \cellcolor{ablationgray}\ab{70.4} & \cellcolor{ablationgray}\ab{61.0} & \cellcolor{ablationgray}\ab{73.3} & \cellcolor{ablationgray}\ab{56.5} \\

\addlinespace[2pt]

\rowcolor{closedsource}
& & GPT 5.2
  & \underline{22.7} & \underline{45.6}
  & \underline{47.0} & \underline{42.6} & \textbf{40.5}
  & 58.8 & 17.9 & \textbf{60.0} & \textbf{63.9} & \textbf{58.6} \\

\rowcolor{closedsource}
\multirow{2}{*}{\rotatebox{90}{\textcolor{textgray}{\scriptsize\textit{oracle}}}}
  & \cellcolor{ablationgray} & \cellcolor{ablationgray}\ab{\small + Image Descriptions}
  & \cellcolor{ablationgray}\ab{37.8} & \cellcolor{ablationgray}\ab{60.8}
  & \cellcolor{ablationgray}\ab{62.9} & \cellcolor{ablationgray}\ab{57.0} & \cellcolor{ablationgray}\ab{56.2}
  & \cellcolor{ablationgray}\ab{70.0} & \cellcolor{ablationgray}\ab{46.8} & \cellcolor{ablationgray}\ab{65.6} & \cellcolor{ablationgray}\ab{69.2} & \cellcolor{ablationgray}\ab{64.3} \\
  & \cellcolor{ablationgray} & \cellcolor{ablationgray}\ab{\small + Image Desc. (w/o img)}
  & \cellcolor{ablationgray}\ab{45.3} & \cellcolor{ablationgray}\ab{66.5}
  & \cellcolor{ablationgray}\ab{69.1} & \cellcolor{ablationgray}\ab{61.2} & \cellcolor{ablationgray}\ab{62.9}
  & \cellcolor{ablationgray}\ab{73.8} & \cellcolor{ablationgray}\ab{56.5} & \cellcolor{ablationgray}\ab{69.3} & \cellcolor{ablationgray}\ab{72.5} & \cellcolor{ablationgray}\ab{68.1} \\

\addlinespace[2pt]

\rowcolor{closedsource}
& & GPT-4o
  & 16.3 & 41.5
  & 42.4 & 41.2 & 30.7
  & \underline{61.8} & \underline{20.5} & 42.3 & 49.8 & 39.2 \\

\rowcolor{closedsource}
\multirow{2}{*}{\rotatebox{90}{\textcolor{textgray}{\scriptsize\textit{oracle}}}}
  & \cellcolor{ablationgray} & \cellcolor{ablationgray}\ab{\small + Image Descriptions}
  & \cellcolor{ablationgray}\ab{23.5} & \cellcolor{ablationgray}\ab{50.5}
  & \cellcolor{ablationgray}\ab{51.9} & \cellcolor{ablationgray}\ab{48.8} & \cellcolor{ablationgray}\ab{43.4}
  & \cellcolor{ablationgray}\ab{65.6} & \cellcolor{ablationgray}\ab{36.4} & \cellcolor{ablationgray}\ab{49.4} & \cellcolor{ablationgray}\ab{58.1} & \cellcolor{ablationgray}\ab{45.8} \\
  & \cellcolor{ablationgray} & \cellcolor{ablationgray}\ab{\small + Image Desc. (w/o img)}
  & \cellcolor{ablationgray}\ab{37.2} & \cellcolor{ablationgray}\ab{62.8}
  & \cellcolor{ablationgray}\ab{65.3} & \cellcolor{ablationgray}\ab{58.8} & \cellcolor{ablationgray}\ab{52.3}
  & \cellcolor{ablationgray}\ab{71.3} & \cellcolor{ablationgray}\ab{56.3} & \cellcolor{ablationgray}\ab{60.7} & \cellcolor{ablationgray}\ab{69.1} & \cellcolor{ablationgray}\ab{57.5} \\

\addlinespace[2pt]

\midrule

\multicolumn{12}{@{}l}{\textit{\small Open source}} \\[1.5pt]

\rowcolor{opensource}
& & Qwen 2.5 7B
  & \textbf{23.0} & 44.2
  & 44.8 & \underline{43.5} & \textbf{41.3}
  & \underline{60.3} & 13.8 & \textbf{58.7} & \textbf{63.3} & \textbf{57.0} \\

\rowcolor{opensource}
\multirow{2}{*}{\rotatebox{90}{\textcolor{textgray}{\scriptsize\textit{oracle}}}}
  & \cellcolor{ablationgray} & \cellcolor{ablationgray}\ab{\small + Image Descriptions}
  & \cellcolor{ablationgray}\ab{30.8} & \cellcolor{ablationgray}\ab{48.9}
  & \cellcolor{ablationgray}\ab{49.2} & \cellcolor{ablationgray}\ab{48.6} & \cellcolor{ablationgray}\ab{46.8}
  & \cellcolor{ablationgray}\ab{67.3} & \cellcolor{ablationgray}\ab{17.9} & \cellcolor{ablationgray}\ab{61.5} & \cellcolor{ablationgray}\ab{63.1} & \cellcolor{ablationgray}\ab{60.9} \\
  & \cellcolor{ablationgray} & \cellcolor{ablationgray}\ab{\small + Image Desc. (w/o img)}
  & \cellcolor{ablationgray}\ab{32.7} & \cellcolor{ablationgray}\ab{53.2}
  & \cellcolor{ablationgray}\ab{55.1} & \cellcolor{ablationgray}\ab{49.3} & \cellcolor{ablationgray}\ab{50.9}
  & \cellcolor{ablationgray}\ab{68.5} & \cellcolor{ablationgray}\ab{29.9} & \cellcolor{ablationgray}\ab{61.1} & \cellcolor{ablationgray}\ab{62.1} & \cellcolor{ablationgray}\ab{60.7} \\

\addlinespace[2pt]

\rowcolor{opensource}
& & LLaVA-OneVision 7B
  & \underline{19.8} & \underline{44.3}
  & \underline{45.2} & 42.9 & 38.5
  & 58.5 & 19.4 & \underline{54.8} & \underline{59.2} & \underline{53.2} \\

\rowcolor{opensource}
\multirow{2}{*}{\rotatebox{90}{\textcolor{textgray}{\scriptsize\textit{oracle}}}}
  & \cellcolor{ablationgray} & \cellcolor{ablationgray}\ab{\small + Image Descriptions}
  & \cellcolor{ablationgray}\ab{27.3} & \cellcolor{ablationgray}\ab{48.5}
  & \cellcolor{ablationgray}\ab{49.4} & \cellcolor{ablationgray}\ab{47.0} & \cellcolor{ablationgray}\ab{44.0}
  & \cellcolor{ablationgray}\ab{65.4} & \cellcolor{ablationgray}\ab{21.1} & \cellcolor{ablationgray}\ab{59.0} & \cellcolor{ablationgray}\ab{62.8} & \cellcolor{ablationgray}\ab{57.4} \\
  & \cellcolor{ablationgray} & \cellcolor{ablationgray}\ab{\small + Image Desc. (w/o img)}
  & \cellcolor{ablationgray}\ab{34.3} & \cellcolor{ablationgray}\ab{53.6}
  & \cellcolor{ablationgray}\ab{55.7} & \cellcolor{ablationgray}\ab{50.1} & \cellcolor{ablationgray}\ab{46.5}
  & \cellcolor{ablationgray}\ab{68.9} & \cellcolor{ablationgray}\ab{27.9} & \cellcolor{ablationgray}\ab{64.1} & \cellcolor{ablationgray}\ab{65.5} & \cellcolor{ablationgray}\ab{63.5} \\

\addlinespace[2pt]

\rowcolor{opensource}
& & Qwen 3 8B
  & 19.6 & 43.6
  & 43.5 & \textbf{45.9} & 32.6
  & \textbf{61.1} & 19.3 & 50.5 & 56.6 & 48.1 \\

\rowcolor{opensource}
\multirow{2}{*}{\rotatebox{90}{\textcolor{textgray}{\scriptsize\textit{oracle}}}}
  & \cellcolor{ablationgray} & \cellcolor{ablationgray}\ab{\small + Image Descriptions}
  & \cellcolor{ablationgray}\ab{28.1} & \cellcolor{ablationgray}\ab{50.0}
  & \cellcolor{ablationgray}\ab{50.7} & \cellcolor{ablationgray}\ab{49.7} & \cellcolor{ablationgray}\ab{43.2}
  & \cellcolor{ablationgray}\ab{67.2} & \cellcolor{ablationgray}\ab{27.1} & \cellcolor{ablationgray}\ab{55.7} & \cellcolor{ablationgray}\ab{59.9} & \cellcolor{ablationgray}\ab{54.0} \\
  & \cellcolor{ablationgray} & \cellcolor{ablationgray}\ab{\small + Image Desc. (w/o img)}
  & \cellcolor{ablationgray}\ab{40.2} & \cellcolor{ablationgray}\ab{60.2}
  & \cellcolor{ablationgray}\ab{62.4} & \cellcolor{ablationgray}\ab{56.1} & \cellcolor{ablationgray}\ab{54.3}
  & \cellcolor{ablationgray}\ab{72.2} & \cellcolor{ablationgray}\ab{41.0} & \cellcolor{ablationgray}\ab{67.4} & \cellcolor{ablationgray}\ab{68.3} & \cellcolor{ablationgray}\ab{67.1} \\

\addlinespace[2pt]

\rowcolor{opensource}
& & Qwen 2.5 3B
  & 18.3 & \textbf{44.6}
  & \textbf{47.3} & 41.9 & 30.9
  & 53.7 & \underline{27.7} & 52.2 & 57.3 & 50.4 \\

\rowcolor{opensource}
\multirow{2}{*}{\rotatebox{90}{\textcolor{textgray}{\scriptsize\textit{oracle}}}}
  & \cellcolor{ablationgray} & \cellcolor{ablationgray}\ab{\small + Image Descriptions}
  & \cellcolor{ablationgray}\ab{27.5} & \cellcolor{ablationgray}\ab{52.2}
  & \cellcolor{ablationgray}\ab{54.0} & \cellcolor{ablationgray}\ab{48.9} & \cellcolor{ablationgray}\ab{46.6}
  & \cellcolor{ablationgray}\ab{63.3} & \cellcolor{ablationgray}\ab{34.7} & \cellcolor{ablationgray}\ab{58.6} & \cellcolor{ablationgray}\ab{61.6} & \cellcolor{ablationgray}\ab{57.6} \\
  & \cellcolor{ablationgray} & \cellcolor{ablationgray}\ab{\small + Image Desc. (w/o img)}
  & \cellcolor{ablationgray}\ab{29.7} & \cellcolor{ablationgray}\ab{56.0}
  & \cellcolor{ablationgray}\ab{57.2} & \cellcolor{ablationgray}\ab{54.3} & \cellcolor{ablationgray}\ab{48.7}
  & \cellcolor{ablationgray}\ab{67.2} & \cellcolor{ablationgray}\ab{43.7} & \cellcolor{ablationgray}\ab{57.0} & \cellcolor{ablationgray}\ab{53.1} & \cellcolor{ablationgray}\ab{58.2} \\

\addlinespace[2pt]

\rowcolor{opensource}
& & GLM-4V 9B
  & 16.7 & 42.5
  & 43.4 & 41.8 & 35.3
  & 59.2 & 23.8 & 44.6 & 47.3 & 43.7 \\

\rowcolor{opensource}
\multirow{2}{*}{\rotatebox{90}{\textcolor{textgray}{\scriptsize\textit{oracle}}}}
  & \cellcolor{ablationgray} & \cellcolor{ablationgray}\ab{\small + Image Descriptions}
  & \cellcolor{ablationgray}\ab{23.1} & \cellcolor{ablationgray}\ab{47.4}
  & \cellcolor{ablationgray}\ab{48.7} & \cellcolor{ablationgray}\ab{45.2} & \cellcolor{ablationgray}\ab{43.3}
  & \cellcolor{ablationgray}\ab{63.5} & \cellcolor{ablationgray}\ab{33.4} & \cellcolor{ablationgray}\ab{45.3} & \cellcolor{ablationgray}\ab{43.7} & \cellcolor{ablationgray}\ab{45.9} \\
  & \cellcolor{ablationgray} & \cellcolor{ablationgray}\ab{\small + Image Desc. (w/o img)}
  & \cellcolor{ablationgray}\ab{31.0} & \cellcolor{ablationgray}\ab{53.9}
  & \cellcolor{ablationgray}\ab{55.8} & \cellcolor{ablationgray}\ab{50.5} & \cellcolor{ablationgray}\ab{48.4}
  & \cellcolor{ablationgray}\ab{66.0} & \cellcolor{ablationgray}\ab{40.2} & \cellcolor{ablationgray}\ab{55.7} & \cellcolor{ablationgray}\ab{56.3} & \cellcolor{ablationgray}\ab{55.5} \\

\addlinespace[2pt]

\rowcolor{opensource}
& & Phi-3V 4.2B
  & 16.1 & 40.2
  & 40.8 & 39.1 & \underline{38.9}
  & 57.1 & 14.7 & 48.8 & 53.2 & 47.3 \\

\rowcolor{opensource}
\multirow{2}{*}{\rotatebox{90}{\textcolor{textgray}{\scriptsize\textit{oracle}}}}
  & \cellcolor{ablationgray} & \cellcolor{ablationgray}\ab{\small + Image Descriptions}
  & \cellcolor{ablationgray}\ab{29.9} & \cellcolor{ablationgray}\ab{49.9}
  & \cellcolor{ablationgray}\ab{50.6} & \cellcolor{ablationgray}\ab{48.7} & \cellcolor{ablationgray}\ab{48.1}
  & \cellcolor{ablationgray}\ab{65.9} & \cellcolor{ablationgray}\ab{23.9} & \cellcolor{ablationgray}\ab{59.8} & \cellcolor{ablationgray}\ab{63.6} & \cellcolor{ablationgray}\ab{58.4} \\
  & \cellcolor{ablationgray} & \cellcolor{ablationgray}\ab{\small + Image Desc. (w/o img)}
  & \cellcolor{ablationgray}\ab{35.8} & \cellcolor{ablationgray}\ab{56.4}
  & \cellcolor{ablationgray}\ab{58.0} & \cellcolor{ablationgray}\ab{53.1} & \cellcolor{ablationgray}\ab{52.5}
  & \cellcolor{ablationgray}\ab{67.3} & \cellcolor{ablationgray}\ab{36.6} & \cellcolor{ablationgray}\ab{65.3} & \cellcolor{ablationgray}\ab{68.5} & \cellcolor{ablationgray}\ab{64.2} \\

\addlinespace[2pt]

\rowcolor{opensource}
& & Aya Vision 8B
  & 15.4 & 37.3
  & 37.9 & 37.5 & 28.1
  & 59.5 & 7.3 & 45.1 & 52.0 & 42.5 \\

\rowcolor{opensource}
\multirow{2}{*}{\rotatebox{90}{\textcolor{textgray}{\scriptsize\textit{oracle}}}}
  & \cellcolor{ablationgray} & \cellcolor{ablationgray}\ab{\small + Image Descriptions}
  & \cellcolor{ablationgray}\ab{26.7} & \cellcolor{ablationgray}\ab{44.3}
  & \cellcolor{ablationgray}\ab{44.9} & \cellcolor{ablationgray}\ab{43.6} & \cellcolor{ablationgray}\ab{39.8}
  & \cellcolor{ablationgray}\ab{64.7} & \cellcolor{ablationgray}\ab{12.1} & \cellcolor{ablationgray}\ab{56.0} & \cellcolor{ablationgray}\ab{57.4} & \cellcolor{ablationgray}\ab{55.5} \\
  & \cellcolor{ablationgray} & \cellcolor{ablationgray}\ab{\small + Image Desc. (w/o img)}
  & \cellcolor{ablationgray}\ab{28.7} & \cellcolor{ablationgray}\ab{48.5}
  & \cellcolor{ablationgray}\ab{49.8} & \cellcolor{ablationgray}\ab{46.4} & \cellcolor{ablationgray}\ab{42.8}
  & \cellcolor{ablationgray}\ab{65.2} & \cellcolor{ablationgray}\ab{23.3} & \cellcolor{ablationgray}\ab{56.9} & \cellcolor{ablationgray}\ab{58.4} & \cellcolor{ablationgray}\ab{56.3} \\

\addlinespace[2pt]

\rowcolor{opensource}
& & Qwen 3 4B
  & 14.9 & 37.7
  & 37.3 & 40.5 & 28.2
  & 59.7 & 12.1 & 41.3 & 43.6 & 40.4 \\

\rowcolor{opensource}
\multirow{2}{*}{\rotatebox{90}{\textcolor{textgray}{\scriptsize\textit{oracle}}}}
  & \cellcolor{ablationgray} & \cellcolor{ablationgray}\ab{\small + Image Descriptions}
  & \cellcolor{ablationgray}\ab{23.0} & \cellcolor{ablationgray}\ab{43.7}
  & \cellcolor{ablationgray}\ab{43.5} & \cellcolor{ablationgray}\ab{45.0} & \cellcolor{ablationgray}\ab{39.8}
  & \cellcolor{ablationgray}\ab{64.4} & \cellcolor{ablationgray}\ab{15.3} & \cellcolor{ablationgray}\ab{51.2} & \cellcolor{ablationgray}\ab{55.7} & \cellcolor{ablationgray}\ab{49.4} \\
  & \cellcolor{ablationgray} & \cellcolor{ablationgray}\ab{\small + Image Desc. (w/o img)}
  & \cellcolor{ablationgray}\ab{30.7} & \cellcolor{ablationgray}\ab{52.9}
  & \cellcolor{ablationgray}\ab{54.7} & \cellcolor{ablationgray}\ab{49.9} & \cellcolor{ablationgray}\ab{47.6}
  & \cellcolor{ablationgray}\ab{67.5} & \cellcolor{ablationgray}\ab{33.7} & \cellcolor{ablationgray}\ab{57.6} & \cellcolor{ablationgray}\ab{58.8} & \cellcolor{ablationgray}\ab{57.1} \\

\addlinespace[2pt]

\rowcolor{opensource}
& & DS-VL2 1B
  & 13.8 & 40.6
  & 41.6 & 39.6 & 34.2
  & 53.8 & \textbf{27.7} & 40.2 & 40.3 & 40.1 \\

\rowcolor{opensource}
\multirow{2}{*}{\rotatebox{90}{\textcolor{textgray}{\scriptsize\textit{oracle}}}}
  & \cellcolor{ablationgray} & \cellcolor{ablationgray}\ab{\small + Image Descriptions}
  & \cellcolor{ablationgray}\ab{19.7} & \cellcolor{ablationgray}\ab{45.4}
  & \cellcolor{ablationgray}\ab{46.2} & \cellcolor{ablationgray}\ab{44.7} & \cellcolor{ablationgray}\ab{38.2}
  & \cellcolor{ablationgray}\ab{60.7} & \cellcolor{ablationgray}\ab{31.1} & \cellcolor{ablationgray}\ab{44.3} & \cellcolor{ablationgray}\ab{40.2} & \cellcolor{ablationgray}\ab{45.6} \\
  & \cellcolor{ablationgray} & \cellcolor{ablationgray}\ab{\small + Image Desc. (w/o img)}
  & \cellcolor{ablationgray}\ab{2.7} & \cellcolor{ablationgray}\ab{26.8}
  & \cellcolor{ablationgray}\ab{27.5} & \cellcolor{ablationgray}\ab{25.2} & \cellcolor{ablationgray}\ab{26.5}
  & \cellcolor{ablationgray}\ab{0.2} & \cellcolor{ablationgray}\ab{20.5} & \cellcolor{ablationgray}\ab{59.8} & \cellcolor{ablationgray}\ab{68.1} & \cellcolor{ablationgray}\ab{57.5} \\

\addlinespace[2pt]

\bottomrule
\end{tabular}%
}
\caption{\normalfont
  Full results for models evaluated in the main paper (Table~1), including per-source and per-label Macro F1 breakdowns.
  Models are ordered by baseline Group Accuracy (descending) within each source group.
  \textbf{Bold} = best within source group; \underline{underline} = second-best.
  Shaded rows show oracle ablations: (1) \textit{+ Image Descriptions} -- image with text description; (2) \textit{+ Image Desc. (w/o img)} -- text description only, no image.
}
\label{tab:app_main12_full}
\end{table*}


\begin{table*}[h]
\centering
\setlength{\tabcolsep}{3pt}
\renewcommand{\arraystretch}{1.05}
\resizebox{\textwidth}{!}{%
\begin{tabular}{@{}
  r @{\hspace{3pt}}
  l
  l
  C C
  @{\hspace{3pt}} !{\color{gray!50}\vrule} @{\hspace{3pt}}
  C C C
  @{\hspace{3pt}} !{\color{gray!50}\vrule} @{\hspace{3pt}}
  C C C C C
  @{}}

\toprule
& & \textbf{Condition}
  & \textbf{Group Acc.} & \textbf{Macro F1}
  & \textbf{Generated} & \textbf{Inpainting} & \textbf{Retrieval}
  & \textbf{Follow} & \textbf{Avoid} & \textbf{Benign\_\textsubscript{All}} & \textbf{Benign\_\textsubscript{Avoid}} & \textbf{Benign\_\textsubscript{Follow}} \\
\midrule

\multicolumn{12}{@{}l}{\textit{\small Open source (remaining models, ordered by baseline Group Accuracy)}} \\[1.5pt]

\rowcolor{opensource}
& & InternVL 2B
  & \textbf{13.4} & 36.2
  & 36.3 & 36.8 & \underline{31.9}
  & \underline{58.9} & 10.2 & \underline{39.4} & \underline{44.0} & 37.7 \\

\rowcolor{opensource}
\multirow{2}{*}{\rotatebox{90}{\textcolor{textgray}{\scriptsize\textit{oracle}}}}
  & \cellcolor{ablationgray} & \cellcolor{ablationgray}\ab{\small + Image Descriptions}
  & \cellcolor{ablationgray}\ab{19.8} & \cellcolor{ablationgray}\ab{39.4}
  & \cellcolor{ablationgray}\ab{40.0} & \cellcolor{ablationgray}\ab{38.5} & \cellcolor{ablationgray}\ab{37.6}
  & \cellcolor{ablationgray}\ab{62.2} & \cellcolor{ablationgray}\ab{8.7} & \cellcolor{ablationgray}\ab{47.3} & \cellcolor{ablationgray}\ab{48.1} & \cellcolor{ablationgray}\ab{47.0} \\
  & \cellcolor{ablationgray} & \cellcolor{ablationgray}\ab{\small + Image Desc. (w/o img)}
  & \cellcolor{ablationgray}\ab{33.7} & \cellcolor{ablationgray}\ab{50.0}
  & \cellcolor{ablationgray}\ab{51.2} & \cellcolor{ablationgray}\ab{46.9} & \cellcolor{ablationgray}\ab{49.0}
  & \cellcolor{ablationgray}\ab{66.5} & \cellcolor{ablationgray}\ab{16.9} & \cellcolor{ablationgray}\ab{66.5} & \cellcolor{ablationgray}\ab{66.6} & \cellcolor{ablationgray}\ab{66.5} \\

\addlinespace[2pt]

\rowcolor{opensource}
& & InternVL 8B
  & \underline{12.9} & \underline{37.7}
  & \underline{37.9} & \textbf{39.3} & 27.1
  & \textbf{59.4} & 16.4 & 37.4 & 43.9 & 35.0 \\

\rowcolor{opensource}
\multirow{2}{*}{\rotatebox{90}{\textcolor{textgray}{\scriptsize\textit{oracle}}}}
  & \cellcolor{ablationgray} & \cellcolor{ablationgray}\ab{\small + Image Descriptions}
  & \cellcolor{ablationgray}\ab{22.3} & \cellcolor{ablationgray}\ab{46.3}
  & \cellcolor{ablationgray}\ab{47.8} & \cellcolor{ablationgray}\ab{44.5} & \cellcolor{ablationgray}\ab{38.1}
  & \cellcolor{ablationgray}\ab{64.3} & \cellcolor{ablationgray}\ab{24.7} & \cellcolor{ablationgray}\ab{49.8} & \cellcolor{ablationgray}\ab{57.1} & \cellcolor{ablationgray}\ab{46.7} \\
  & \cellcolor{ablationgray} & \cellcolor{ablationgray}\ab{\small + Image Desc. (w/o img)}
  & \cellcolor{ablationgray}\ab{35.3} & \cellcolor{ablationgray}\ab{57.6}
  & \cellcolor{ablationgray}\ab{60.0} & \cellcolor{ablationgray}\ab{53.5} & \cellcolor{ablationgray}\ab{51.5}
  & \cellcolor{ablationgray}\ab{70.0} & \cellcolor{ablationgray}\ab{41.1} & \cellcolor{ablationgray}\ab{61.8} & \cellcolor{ablationgray}\ab{64.4} & \cellcolor{ablationgray}\ab{60.8} \\

\addlinespace[2pt]

\rowcolor{opensource}
& & Gemma 3 4B
  & 11.7 & 34.6
  & 34.7 & 35.1 & 31.0
  & 58.7 & 10.5 & 34.7 & 39.6 & 32.9 \\

\rowcolor{opensource}
\multirow{2}{*}{\rotatebox{90}{\textcolor{textgray}{\scriptsize\textit{oracle}}}}
  & \cellcolor{ablationgray} & \cellcolor{ablationgray}\ab{\small + Image Descriptions}
  & \cellcolor{ablationgray}\ab{14.0} & \cellcolor{ablationgray}\ab{36.6}
  & \cellcolor{ablationgray}\ab{37.3} & \cellcolor{ablationgray}\ab{35.5} & \cellcolor{ablationgray}\ab{35.5}
  & \cellcolor{ablationgray}\ab{60.1} & \cellcolor{ablationgray}\ab{11.8} & \cellcolor{ablationgray}\ab{37.9} & \cellcolor{ablationgray}\ab{43.1} & \cellcolor{ablationgray}\ab{35.9} \\
  & \cellcolor{ablationgray} & \cellcolor{ablationgray}\ab{\small + Image Desc. (w/o img)}
  & \cellcolor{ablationgray}\ab{25.8} & \cellcolor{ablationgray}\ab{48.1}
  & \cellcolor{ablationgray}\ab{49.2} & \cellcolor{ablationgray}\ab{46.5} & \cellcolor{ablationgray}\ab{42.7}
  & \cellcolor{ablationgray}\ab{64.4} & \cellcolor{ablationgray}\ab{25.1} & \cellcolor{ablationgray}\ab{54.7} & \cellcolor{ablationgray}\ab{59.8} & \cellcolor{ablationgray}\ab{52.7} \\

\addlinespace[2pt]

\rowcolor{opensource}
& & LLaVA-NeXT 1.6 7B
  & 10.8 & \textbf{38.5}
  & \textbf{40.8} & \underline{37.1} & 22.8
  & 42.6 & \textbf{33.9} & 39.1 & 36.2 & \underline{39.9} \\

\rowcolor{opensource}
\multirow{2}{*}{\rotatebox{90}{\textcolor{textgray}{\scriptsize\textit{oracle}}}}
  & \cellcolor{ablationgray} & \cellcolor{ablationgray}\ab{\small + Image Descriptions}
  & \cellcolor{ablationgray}\ab{20.5} & \cellcolor{ablationgray}\ab{50.5}
  & \cellcolor{ablationgray}\ab{52.3} & \cellcolor{ablationgray}\ab{48.5} & \cellcolor{ablationgray}\ab{39.2}
  & \cellcolor{ablationgray}\ab{65.8} & \cellcolor{ablationgray}\ab{42.4} & \cellcolor{ablationgray}\ab{43.2} & \cellcolor{ablationgray}\ab{42.1} & \cellcolor{ablationgray}\ab{43.6} \\
  & \cellcolor{ablationgray} & \cellcolor{ablationgray}\ab{\small + Image Desc. (w/o img)}
  & \cellcolor{ablationgray}\ab{26.6} & \cellcolor{ablationgray}\ab{55.4}
  & \cellcolor{ablationgray}\ab{56.4} & \cellcolor{ablationgray}\ab{54.9} & \cellcolor{ablationgray}\ab{45.7}
  & \cellcolor{ablationgray}\ab{66.1} & \cellcolor{ablationgray}\ab{44.8} & \cellcolor{ablationgray}\ab{55.1} & \cellcolor{ablationgray}\ab{57.8} & \cellcolor{ablationgray}\ab{54.2} \\

\addlinespace[2pt]

\rowcolor{opensource}
& & Molmo 2 8B
  & 5.0 & 31.5
  & 32.0 & 32.1 & 24.3
  & 56.4 & 31.9 & 6.1 & 10.3 & 4.6 \\

\rowcolor{opensource}
\multirow{2}{*}{\rotatebox{90}{\textcolor{textgray}{\scriptsize\textit{oracle}}}}
  & \cellcolor{ablationgray} & \cellcolor{ablationgray}\ab{\small + Image Descriptions}
  & \cellcolor{ablationgray}\ab{22.4} & \cellcolor{ablationgray}\ab{47.2}
  & \cellcolor{ablationgray}\ab{48.8} & \cellcolor{ablationgray}\ab{45.8} & \cellcolor{ablationgray}\ab{37.3}
  & \cellcolor{ablationgray}\ab{65.8} & \cellcolor{ablationgray}\ab{32.8} & \cellcolor{ablationgray}\ab{43.2} & \cellcolor{ablationgray}\ab{45.2} & \cellcolor{ablationgray}\ab{42.5} \\
  & \cellcolor{ablationgray} & \cellcolor{ablationgray}\ab{\small + Image Desc. (w/o img)}
  & \cellcolor{ablationgray}\ab{30.1} & \cellcolor{ablationgray}\ab{52.8}
  & \cellcolor{ablationgray}\ab{54.7} & \cellcolor{ablationgray}\ab{49.3} & \cellcolor{ablationgray}\ab{48.5}
  & \cellcolor{ablationgray}\ab{67.5} & \cellcolor{ablationgray}\ab{35.6} & \cellcolor{ablationgray}\ab{55.4} & \cellcolor{ablationgray}\ab{61.2} & \cellcolor{ablationgray}\ab{53.2} \\

\addlinespace[2pt]

\rowcolor{opensource}
& & BLIP-2 2.7B
  & 3.8 & 26.3
  & 25.7 & 25.5 & \textbf{33.1}
  & 7.1 & 13.9 & \textbf{57.9} & \textbf{63.5} & \textbf{56.3} \\

\rowcolor{opensource}
\multirow{2}{*}{\rotatebox{90}{\textcolor{textgray}{\scriptsize\textit{oracle}}}}
  & \cellcolor{ablationgray} & \cellcolor{ablationgray}\ab{\small + Image Descriptions}
  & \cellcolor{ablationgray}\ab{4.2} & \cellcolor{ablationgray}\ab{26.4}
  & \cellcolor{ablationgray}\ab{25.7} & \cellcolor{ablationgray}\ab{25.4} & \cellcolor{ablationgray}\ab{36.4}
  & \cellcolor{ablationgray}\ab{11.9} & \cellcolor{ablationgray}\ab{9.3} & \cellcolor{ablationgray}\ab{58.0} & \cellcolor{ablationgray}\ab{62.7} & \cellcolor{ablationgray}\ab{56.6} \\
  & \cellcolor{ablationgray} & \cellcolor{ablationgray}\ab{\small + Image Desc. (w/o img)}
  & \cellcolor{ablationgray}\ab{1.9} & \cellcolor{ablationgray}\ab{25.7}
  & \cellcolor{ablationgray}\ab{26.0} & \cellcolor{ablationgray}\ab{25.3} & \cellcolor{ablationgray}\ab{24.0}
  & \cellcolor{ablationgray}\ab{2.2} & \cellcolor{ablationgray}\ab{20.8} & \cellcolor{ablationgray}\ab{54.1} & \cellcolor{ablationgray}\ab{58.5} & \cellcolor{ablationgray}\ab{52.8} \\

\addlinespace[2pt]

\rowcolor{opensource}
& & PaliGemma 3B
  & 2.0 & 19.6
  & 19.3 & 18.5 & 21.9
  & 0.0 & 31.9 & 27.8 & 25.5 & 28.4 \\

\rowcolor{opensource}
\multirow{2}{*}{\rotatebox{90}{\textcolor{textgray}{\scriptsize\textit{oracle}}}}
  & \cellcolor{ablationgray} & \cellcolor{ablationgray}\ab{\small + Image Descriptions}
  & \cellcolor{ablationgray}\ab{1.4} & \cellcolor{ablationgray}\ab{20.4}
  & \cellcolor{ablationgray}\ab{19.8} & \cellcolor{ablationgray}\ab{20.1} & \cellcolor{ablationgray}\ab{21.1}
  & \cellcolor{ablationgray}\ab{0.0} & \cellcolor{ablationgray}\ab{32.3} & \cellcolor{ablationgray}\ab{29.3} & \cellcolor{ablationgray}\ab{25.1} & \cellcolor{ablationgray}\ab{30.5} \\
  & \cellcolor{ablationgray} & \cellcolor{ablationgray}\ab{\small + Image Desc. (w/o img)}
  & \cellcolor{ablationgray}\ab{0.0} & \cellcolor{ablationgray}\ab{0.0}
  & \cellcolor{ablationgray}\ab{0.0} & \cellcolor{ablationgray}\ab{0.0} & \cellcolor{ablationgray}\ab{0.0}
  & \cellcolor{ablationgray}\ab{0.0} & \cellcolor{ablationgray}\ab{0.0} & \cellcolor{ablationgray}\ab{0.0} & \cellcolor{ablationgray}\ab{0.0} & \cellcolor{ablationgray}\ab{0.0} \\

\addlinespace[2pt]

\rowcolor{opensource}
& & LLaVA 1.5 7B
  & 1.5 & 22.3
  & 23.3 & 20.8 & 17.9
  & 34.7 & \underline{32.4} & 0.0 & 0.0 & 0.0 \\

\rowcolor{opensource}
\multirow{2}{*}{\rotatebox{90}{\textcolor{textgray}{\scriptsize\textit{oracle}}}}
  & \cellcolor{ablationgray} & \cellcolor{ablationgray}\ab{\small + Image Descriptions}
  & \cellcolor{ablationgray}\ab{5.3} & \cellcolor{ablationgray}\ab{36.5}
  & \cellcolor{ablationgray}\ab{37.3} & \cellcolor{ablationgray}\ab{35.3} & \cellcolor{ablationgray}\ab{31.1}
  & \cellcolor{ablationgray}\ab{65.0} & \cellcolor{ablationgray}\ab{39.7} & \cellcolor{ablationgray}\ab{4.7} & \cellcolor{ablationgray}\ab{3.0} & \cellcolor{ablationgray}\ab{5.3} \\
  & \cellcolor{ablationgray} & \cellcolor{ablationgray}\ab{\small + Image Desc. (w/o img)}
  & \cellcolor{ablationgray}\ab{5.0} & \cellcolor{ablationgray}\ab{36.3}
  & \cellcolor{ablationgray}\ab{36.9} & \cellcolor{ablationgray}\ab{35.6} & \cellcolor{ablationgray}\ab{31.1}
  & \cellcolor{ablationgray}\ab{66.3} & \cellcolor{ablationgray}\ab{40.2} & \cellcolor{ablationgray}\ab{2.3} & \cellcolor{ablationgray}\ab{1.9} & \cellcolor{ablationgray}\ab{2.5} \\

\addlinespace[2pt]

\rowcolor{opensource}
& & PaliGemma 2 3B
  & 0.0 & 18.8
  & 18.6 & 18.5 & 20.8
  & 56.3 & 0.0 & 0.0 & 0.0 & 0.0 \\

\rowcolor{opensource}
\multirow{2}{*}{\rotatebox{90}{\textcolor{textgray}{\scriptsize\textit{oracle}}}}
  & \cellcolor{ablationgray} & \cellcolor{ablationgray}\ab{\small + Image Descriptions}
  & \cellcolor{ablationgray}\ab{0.0} & \cellcolor{ablationgray}\ab{18.6}
  & \cellcolor{ablationgray}\ab{18.5} & \cellcolor{ablationgray}\ab{18.4} & \cellcolor{ablationgray}\ab{20.0}
  & \cellcolor{ablationgray}\ab{56.2} & \cellcolor{ablationgray}\ab{0.0} & \cellcolor{ablationgray}\ab{0.0} & \cellcolor{ablationgray}\ab{0.0} & \cellcolor{ablationgray}\ab{0.0} \\
  & \cellcolor{ablationgray} & \cellcolor{ablationgray}\ab{\small + Image Desc. (w/o img)}
  & \cellcolor{ablationgray}\ab{0.0} & \cellcolor{ablationgray}\ab{0.0}
  & \cellcolor{ablationgray}\ab{0.0} & \cellcolor{ablationgray}\ab{0.0} & \cellcolor{ablationgray}\ab{0.0}
  & \cellcolor{ablationgray}\ab{0.0} & \cellcolor{ablationgray}\ab{0.0} & \cellcolor{ablationgray}\ab{0.0} & \cellcolor{ablationgray}\ab{0.0} & \cellcolor{ablationgray}\ab{0.0} \\

\addlinespace[2pt]

\bottomrule
\end{tabular}%
}
\caption{\normalfont
  Full results for the remaining 9 open-source models not included in Table~1, ordered by baseline Group Accuracy (descending).
  \textbf{Bold} = best within this group.
  Shaded rows show oracle ablations: (1) \textit{+ Image Descriptions} -- image with text description; (2) \textit{+ Image Desc. (w/o img)} -- text description only, no image. Note the large gap between baseline and ablation for Molmo 2 8B (5.0 $\to$ 30.1) group accuracy with image descriptions, indicating especially poor vision-language alignment.
}
\label{tab:app_remaining9_full}
\end{table*}

\subsection{Effect of Model Size.}
\label{app:ssec:model_size_effect}
Model size shows weak positive Pearson correlations with performance (F1: $r=0.29$; group accuracy: $r=0.26$), but scaling effects vary substantially across families. Qwen 3 exhibits the most robust scaling, with both F1 and group accuracy improving consistently from 4B to 8B (+5.9 and +4.7 percentage points, respectively). Other families show inconsistent patterns: Qwen 2.5 7B improves in group accuracy (23.0\% vs.\ 18.3\%) but not F1 (44.2\% vs.\ 44.6\%) over Qwen 2.5 3B; and InternVL 8B shows minimal or negative change relative to the InternVL 2B variant on both metrics (37.7\% vs.\ 36.2\% F1; 12.9\% vs.\ 13.4\% group accuracy).

\begin{figure}[t]
    \centering
    \includegraphics[width=0.6\columnwidth]{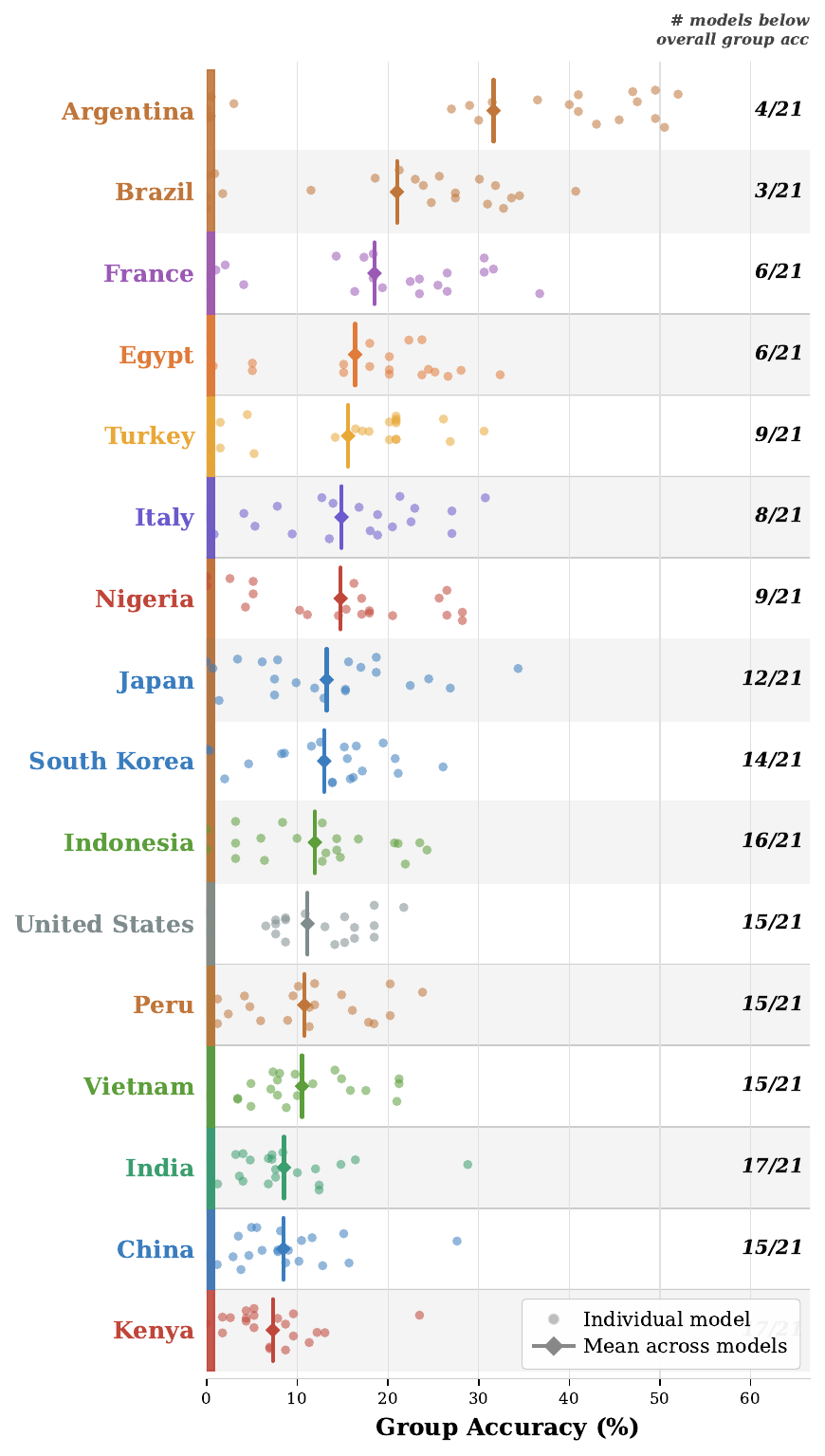}
    \caption{Group accuracy of 21 vision-language models (VLMs) across 16 countries. Each dot represents an individual model's group accuracy (\%) for a given country, and the diamond marker shows the mean across all models. Countries are ordered from lowest (bottom) to highest (top) mean group accuracy. The fraction on the right (e.g., 17/21) indicates how many models score below their own overall group accuracy. For example, a model with an overall mean group accuracy of 15\% that scores 8\% on India would count toward this fraction. Higher fraction indicates that the country is consistently harder for most models relative to their average performance across all countries. Countries are color-coded by geographic region.}
    \label{app:fig:across_countries}
\end{figure}

\subsection{Per Country Performance}
\label{app:ssec:per_country}
Countries in the Global South and non-Western regions prove consistently harder for most models. Kenya and India show the highest fractions of underperforming models (17 out of 21), followed closely by Indonesia (16 out of 21) and a four-way tie among China, Peru, the United States, and Vietnam (15 out of 21), suggesting that VLMs struggle disproportionately with visual norms from these regions relative to their own overall performance. In contrast, Brazil, Argentina, Egypt, and France see far fewer models underperforming (3--6 out of 21), indicating more consistent recognition of visual norms from these countries. This asymmetry points to a systematic gap in how well current VLMs generalize across cultures, likely reflecting imbalances in the cultural composition of their training data. Figure \ref{app:fig:across_countries} shows the performance of 21 VLMs across 16 countries.

\clearpage

\section{Improving Cultural Visual Reasoning}
\label{app:sec:improving}

\subsection{\trainName Construction Additional Details} 
\label{app:ssec:train_construction_details}
We construct \trainName from CANDLE \citep{nguyen2023extracting}. Norm de-contamination between training and benchmark data is crucial, so we perform explicit norm-level semantic deduplication between CANDLE and \benchmarkName's evaluation set before constructing \trainName. For each of the 16 countries, we provide GPT-4o with each CANDLE candidate norm and the full set of test norms from that country (23--93 norms per country; mean 57.5, std.\ 22.4), and prompt it to identify semantic matches regardless of surface phrasing, discarding any candidate flagged as a match. This involves 62,323 GPT-4o calls across all countries. Please find the prompt used in Figure \ref{fig:prompt_decontamination}. To validate the efficacy of GPT-4o filtering, one author manually validated 50 randomly sampled judgments: 48 were in perfect agreement with the model, and the remaining 2 were false positives (over-flagged as matches when they were not), with no false negatives identified---indicating the filter is conservative in the direction of removing more candidate overlap rather than less. 

On the resulting decontaminated set, we apply the same pipeline as \S\ref{sec:benchmark}: structuring norms into \textit{norm--contrast} pairs, extracting metadata, and filtering for visual detectability. To prioritize data scale over source diversity, we rely solely on text-to-image generation. After filtering CANDLE across 16 countries for visual detectability, we retain 15,331 cultural descriptors of highly visual social behaviors, each tagged as Follow or Avoid.
For each norm, we generate 10 T2I prompts for both the Follow/Avoid and its corresponding Benign counterpart (Benign\_Follow or Benign\_Avoid), then generate 10 images per prompt using \texttt{Stable Diffusion 3 (SD3)} \citep{esser2024scaling}, yielding 100 images per norm variant and 100 for the corresponding benign version.
Next, we score each norm image against both its own prompt and its benign counterpart using VQA (as in \S\ref{ssec:image_curation}), and vice versa for the benign images, yielding four cross-evaluated scores per image pair. We then compute a discriminability score per pair: $= (norm\_image\_on\_norm\_prompt + benign\_image\_on\_benign\_prompt) --
- (benign\_image\_on\_norm\_prompt + norm\_image\_on\_benign\_prompt)$. This rewards pairs where each image aligns strongly with its intended prompt while remaining semantically distant from the other. We retain only pairs exceeding a discriminability threshold of 0.8, and select the top $n \in \{10, 20\}$ pairs per norm, treating $n$ as a control variable in our experiments. Through this rigorous filtering, we ensure data quality of \trainName. We also run a targeted training experiment using only the \benchmarkName norms, generating a different training set using it, and measure performance difference only to find only marginal improvements (later in \S\ref{app:ssec:train-test-run}).

\begin{figure}[!htbp]
\begin{tcolorbox}[
  colback=gray!5,
  colframe=gray!75!black,
  title={\textbf{\trainName-\benchmarkName norm decontamination}},
  fonttitle=\bfseries,
  coltitle=white,
  colbacktitle=gray!75!black,
]
\small
\begin{Verbatim}[breaklines=true, breakanywhere=true]
System Prompt: "You are a helpful assistant that analyzes whether social norms are semantically similar. You will be given a candidate norm and a list of reference norms. Your task is to identify if the candidate norm semantically matches any of the reference norms. Return a JSON with three fields: 1. "is_match": true if there's a semantic match, false otherwise 2. "matching_norms": list of indices of matching norms (empty if no match) 3. "explanation": brief explanation of your reasoning"

User Prompt: "Candidate norm: {train_norm}. Reference norms: {formatted_test_norms}. Determine if the candidate norm semantically matches any of the reference norms. Consider that norms can be phrased differently but have the same meaning. Focus on the core social rule rather than superficial wording differences. Return your analysis in JSON format as specified."
\end{Verbatim}
\normalsize
\end{tcolorbox}
\caption{Prompt used for excluding \benchmarkName norms from \trainName}
\label{fig:prompt_decontamination}
\end{figure}

\subsection{Experimental Setup}
\label{app:ssec:exp_setup_details}
We experiment with three label distribution configurations: \textit{balanced} (25\% each of \textit{Follow}, \textit{Avoid}, \textit{Benign\textsubscript{Follow}}, \textit{Benign\textsubscript{Avoid}}), \textit{follow-heavy} (35\% \textit{Follow}, 35\% \textit{Benign\textsubscript{Follow}}, 15\% \textit{Avoid}, 15\% \textit{Benign\textsubscript{Avoid}}), and \textit{avoid-heavy} (the inverse).  We additionally vary data scale by selecting top $n = {10, 20}$ images per cultural norm ranked by VQA score (out of the generated 100 images), yielding 32,144 and 64,288 training examples respectively (32k and 64k). We finetune Qwen3 VL 4B and 8B models on both 32k and 64k. We also experiment with Qwen2.5 VL 7B model on 32k, and see minimal gains. 

While we agree that these experiments are limited to the Qwen model family, there are meaningful differences between Qwen 2.5 VL and Qwen 3 VL models. Qwen2.5-VL model framework uses a Vision Transformer (ViT) vision encoder trained from scratch on DataComp with a Qwen 2.5 LLM as the language decoder, whereas Qwen 3 VL switches to SigLIP-2 vision encoder pretrained on WebLI data and Qwen 3 LLM as the decoder. These differ not only in architecture but also in vision encoder training data, and we observe consistent gains across all variants (Appendix \ref{app:ssec:finetuning_results_detailed}), suggesting the training signal in \trainName is not specific to a single architecture. 

\subsection{Full Results discussing effects of label distributions and data scales.}
\label{app:ssec:finetuning_results_detailed}
\begin{figure}[t]
    \centering
    \includegraphics[width=\columnwidth]{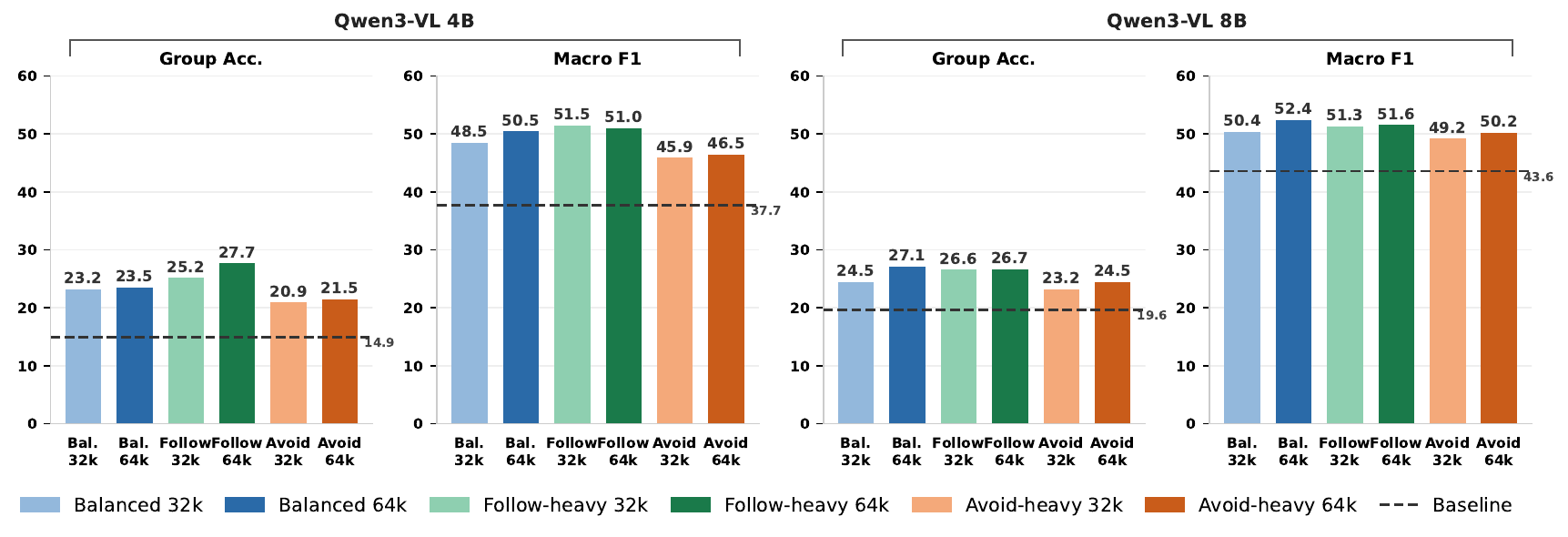}
     \caption{Finetuning results for Qwen3 VL 4B and 8B models across label distribution configurations (balanced, follow-heavy, avoid-heavy) and data scales (32k and 64k)} 
    \label{app:fig:finetuning_results}
\end{figure}

\begin{figure}[t]
    \centering
    \includegraphics[width=0.5\linewidth]{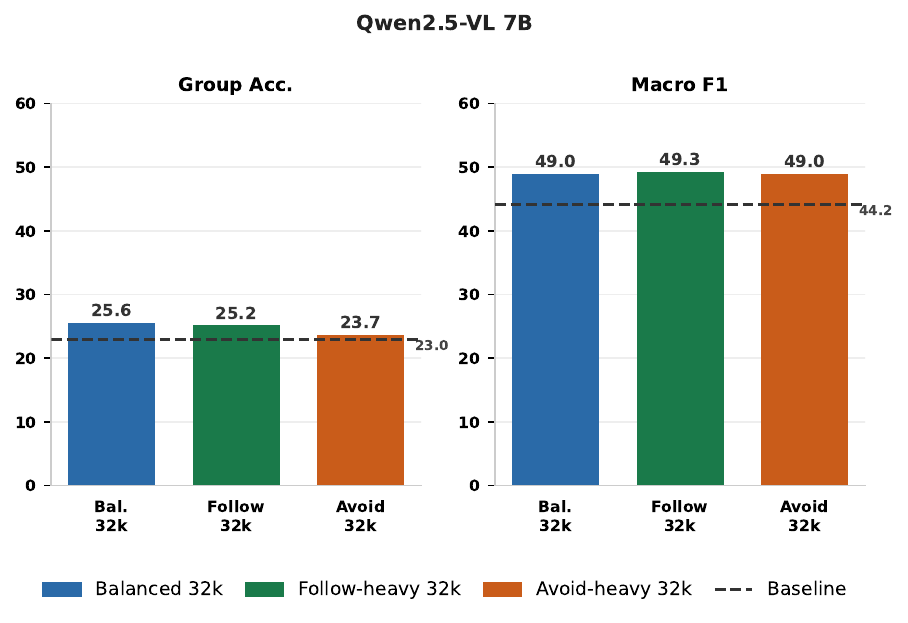}
    \caption{Finetuning results for Qwen 2.5 VL 7B model across label distribution configurations (balanced, follow-heavy, avoid-heavy) and 32k data size}
    \label{app:fig:finetuning_results_qwen25}
\end{figure}

Finetuning Qwen3 VL 4B and 8B models on balanced, follow-heavy and avoid-heavy configurations and data scales of \trainName consistently helps across the board. \textit{Balanced} and \textit{follow-heavy} training perform competitively across all label types, while \textit{avoid-heavy} configuration yields relatively lower gains likely due to the test set being skewed toward the \textit{Follow} label. Scaling from $32k$ to $64k$ yields almost consistent but modest gains across all configurations and both models, (group accuracy: +0.1-2.6, F1: +0.3-2.0, except \textit{follow-heavy} $64k$ which drops by -0.5). Absolute group accuracy still remains low across the board, indicating that meaningful gains likely require training objectives that go beyond label prediction, perhaps with additional supervision such as bounding boxes to better localize cultural nuances, or alternative contrastive training objectives such as CLIP \citep{radford2021learning}. See Figure \ref{app:fig:finetuning_results} for complete results. We also initially experiment with Qwen 2.5VL 7B (see Figure \ref{app:fig:finetuning_results_qwen25}) and see moderate gains.

\definecolor{finegray}{RGB}{220, 240, 220}

\subsection{Probing for training-test leakage}
\label{app:ssec:train-test-run}
We additionally ran a preliminary targeted experiment to specifically probe whether leakage is a meaningful concern: we constructed a separate training subset using only the test set norms directly (i.e., explicitly training on the same norms as \benchmarkName), generated images using the same SD3 pipeline, and finetuned both Qwen3-VL 4B and 8B on this set. We observe only marginal improvements on balanced label distribution and 32k datasize: Qwen3-VL 4B improves from $23.2$ group accuracy (F1: $48.5$) to $23.9$ (F1: $52.0$), and Qwen3-VL 8B from $24.5$ (F1: $50.4$) to $26.7$ (F1: $53.4$), suggesting that the lack of norm-level information is not the bottleneck for the current finetuning setup, but rather that meaningful gains likely require richer training signals, better vision-language alignment for cultural content, or architectures better suited to perceiving and reasoning about cultural behaviors.

\subsection{Improvement Across Countries and Semantic Cultural Nuances}
\label{app:ssec:improvements_across_countries}
\begin{figure}[htbp]
  \centering
  \includegraphics[width=0.48\linewidth]{appendix_figures/qwen3_4b_img20_70_30_group_acc_delta.pdf}
  \hfill
  \includegraphics[width=0.48\linewidth]{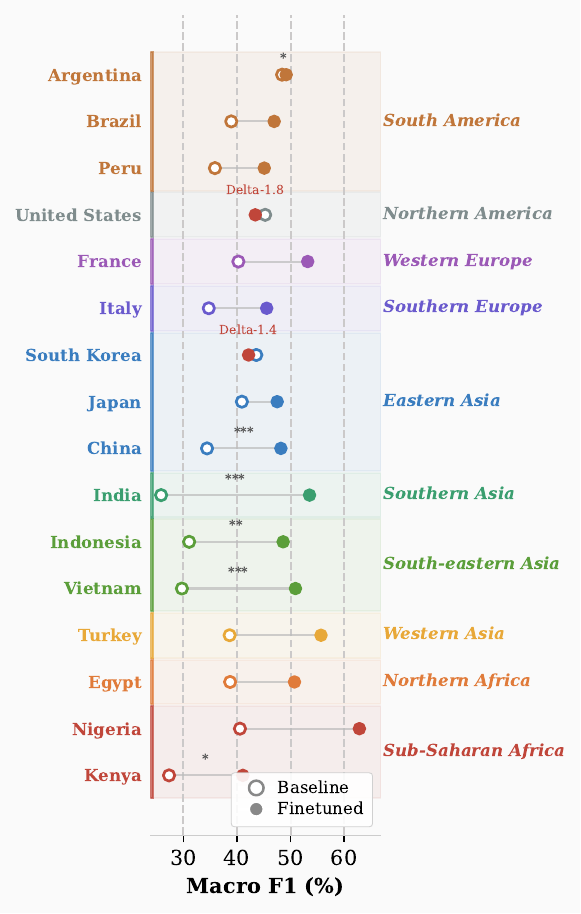}
  \caption{Country-wise performance of finetuned Qwen3VL 4B model, using group accuracy and single image classification F1 score. Results for follow-heavy 64k finetuned Qwen3VL 4B on \trainName}
  \label{app:fig:finetuned_qwen4b_across_countries}
\end{figure}

\begin{figure}[htbp]
  \centering
  \includegraphics[width=0.48\linewidth]{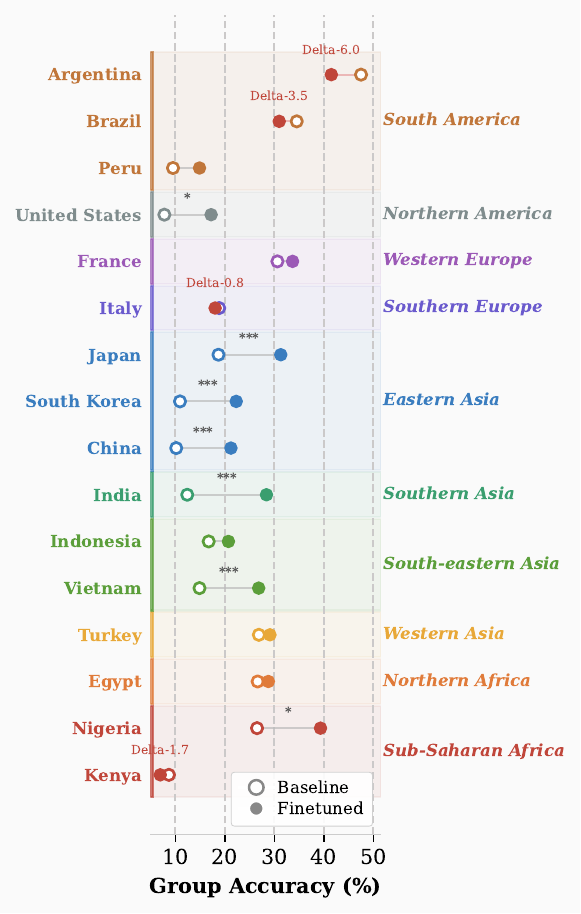}
  \hfill
  \includegraphics[width=0.48\linewidth]{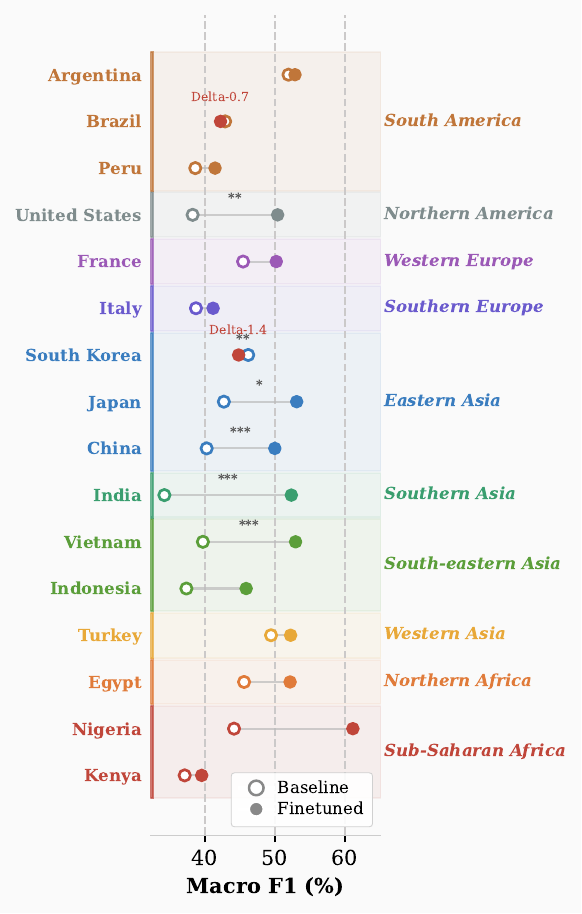}
  \caption{Country-wise performance of finetuned Qwen3VL 8B model, using group accuracy and single image classification F1 score. Results for follow-heavy 64k finetuned Qwen3VL 8B on \trainName}
  \label{app:fig:finetuned_qwen8b_across_countries}
\end{figure}

\begin{figure}
    \centering
    \includegraphics[width=\linewidth]{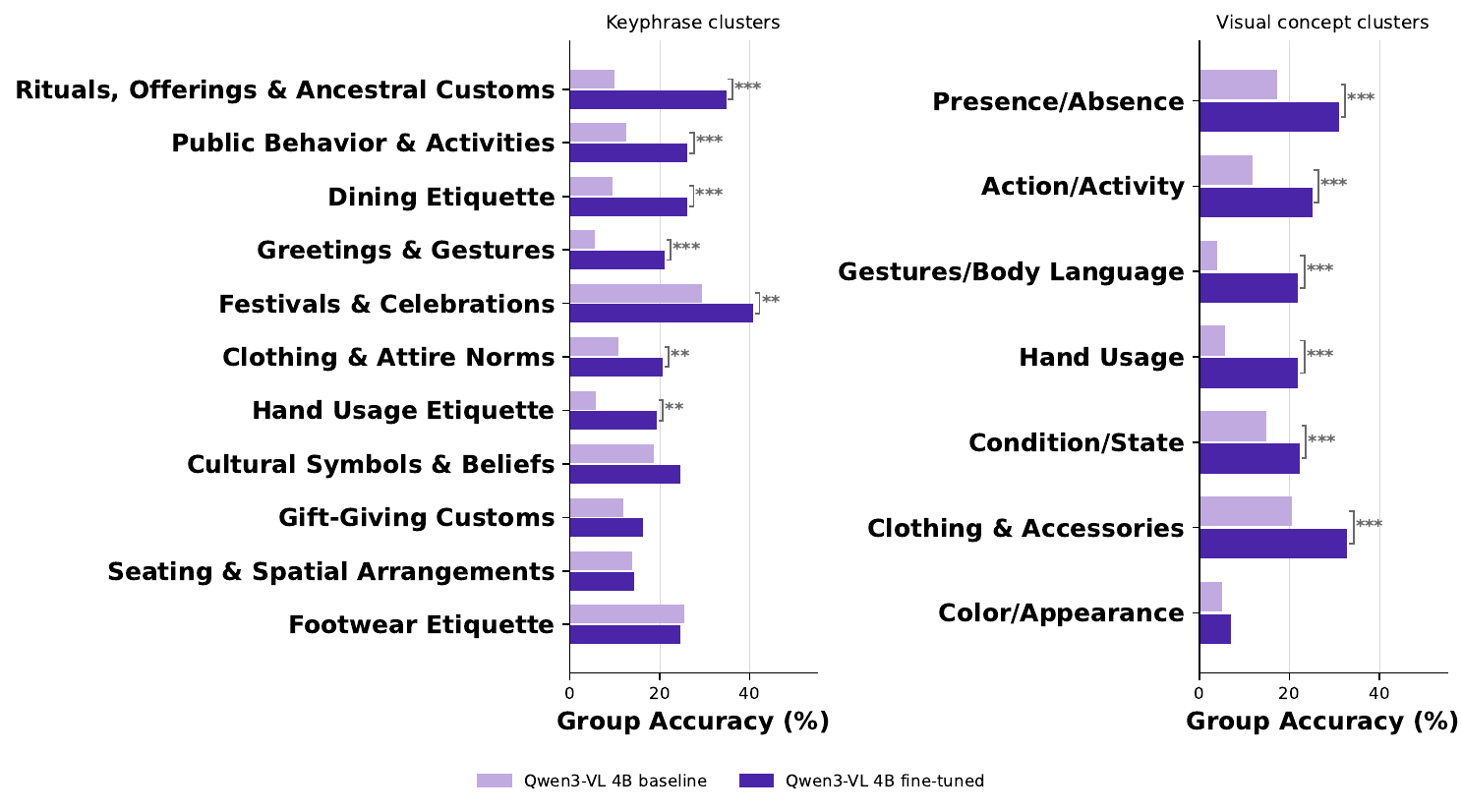}
    \caption{Improvement of finetuned Qwen3 VL 4B (follow-heavy 64k) across cultural norm domains and visual contrast types. Improvements measured using group accuracy.}
    \label{app:fig:finetuned_4b_group_acc_semantic}
\end{figure}

\begin{figure}
    \centering
    \includegraphics[width=\linewidth]{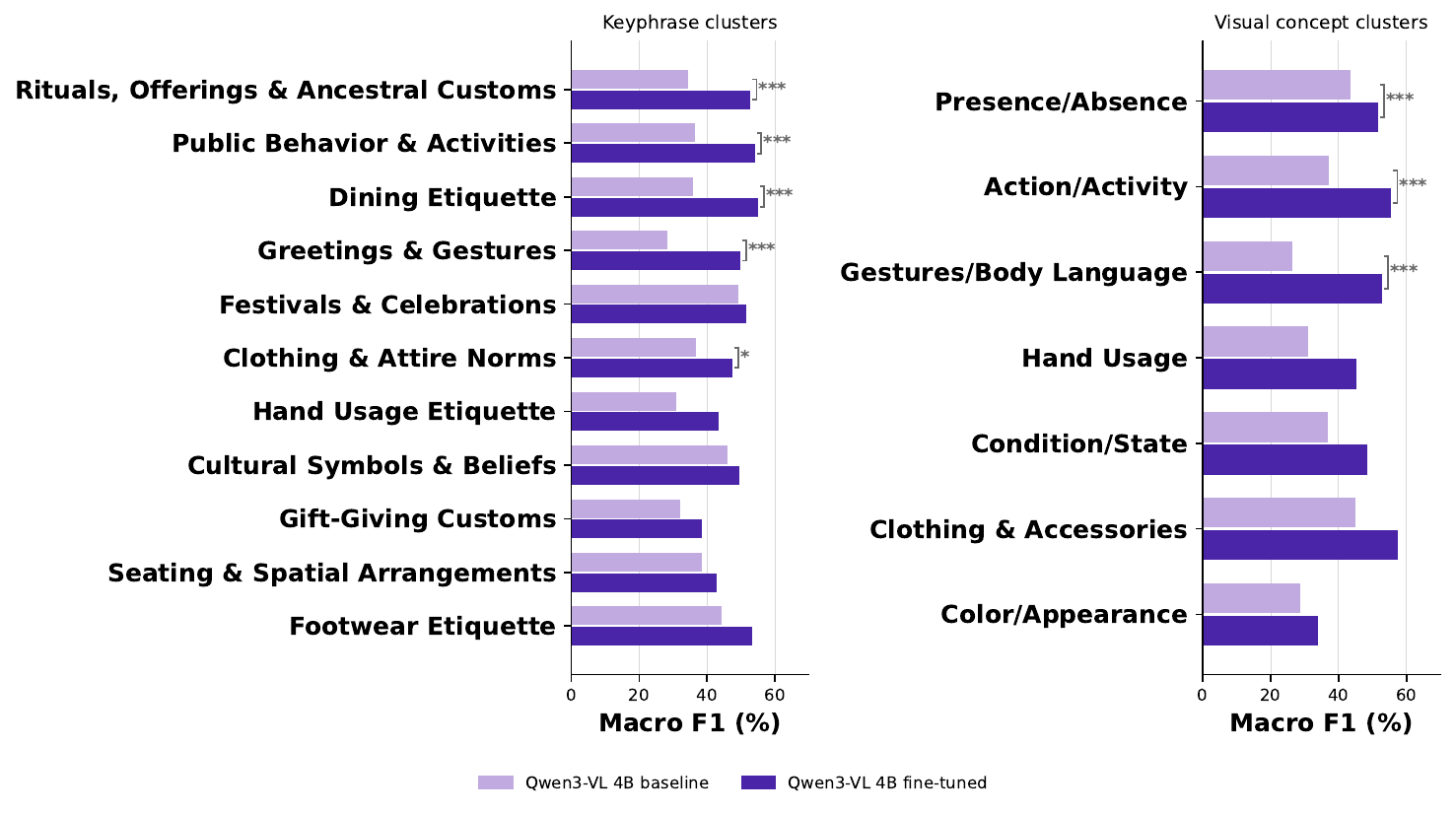}
    \caption{Improvement of finetuned Qwen3 VL 4B (follow-heavy 64k) across cultural norm domains and visual contrast types. Improvements measured using single image classification F1.}
    \label{app:fig:finetuned_4b_f1_semantic}
\end{figure}

\begin{figure}
    \centering
    \includegraphics[width=\linewidth]{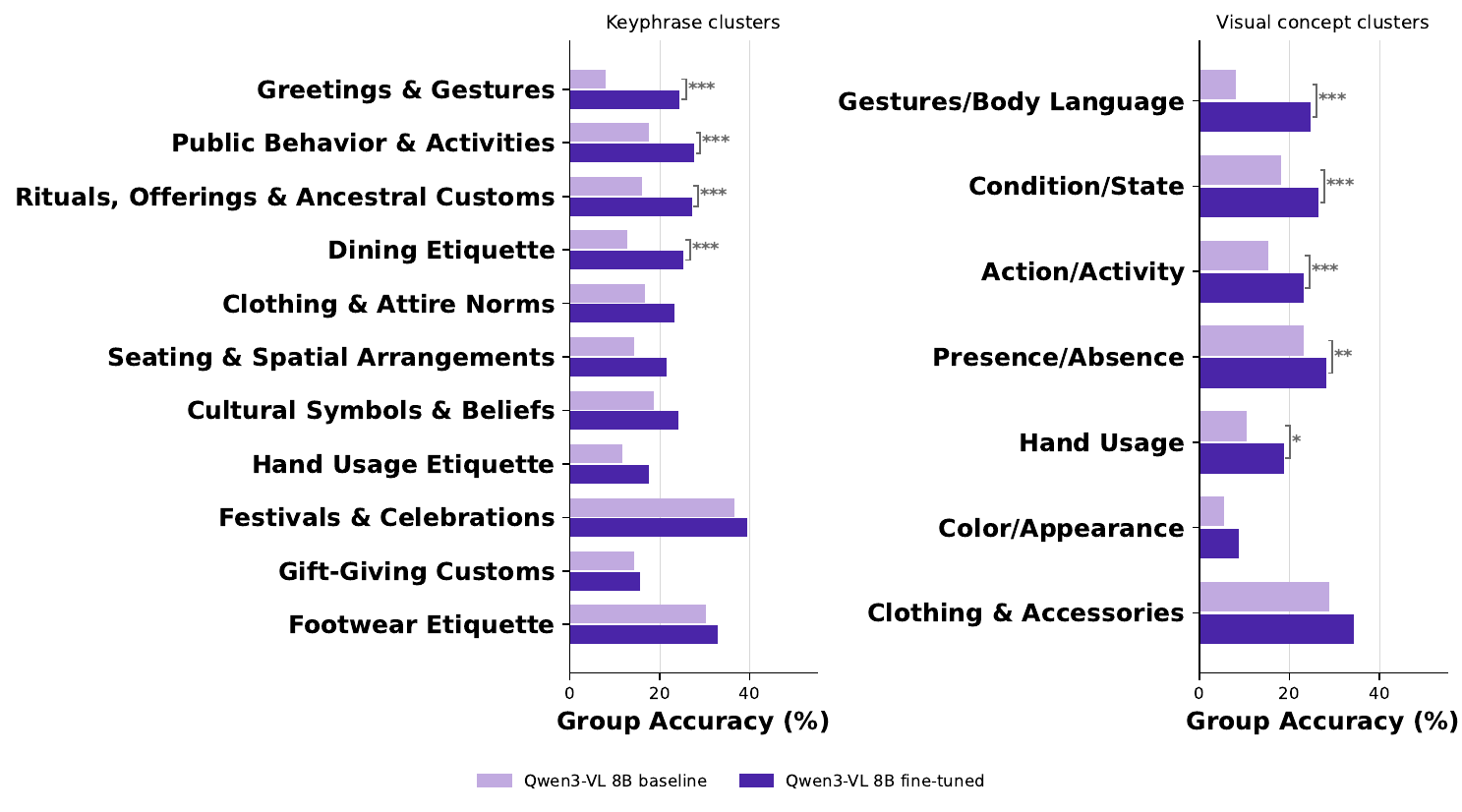}
    \caption{Improvement of finetuned Qwen3 VL 8B (follow-heavy 64k) across cultural norm domains and visual contrast types. Improvements measured using group accuracy.}
    \label{app:fig:finetuned_8b_group_acc_semantic}
\end{figure}

\begin{figure}
    \centering
    \includegraphics[width=\linewidth]{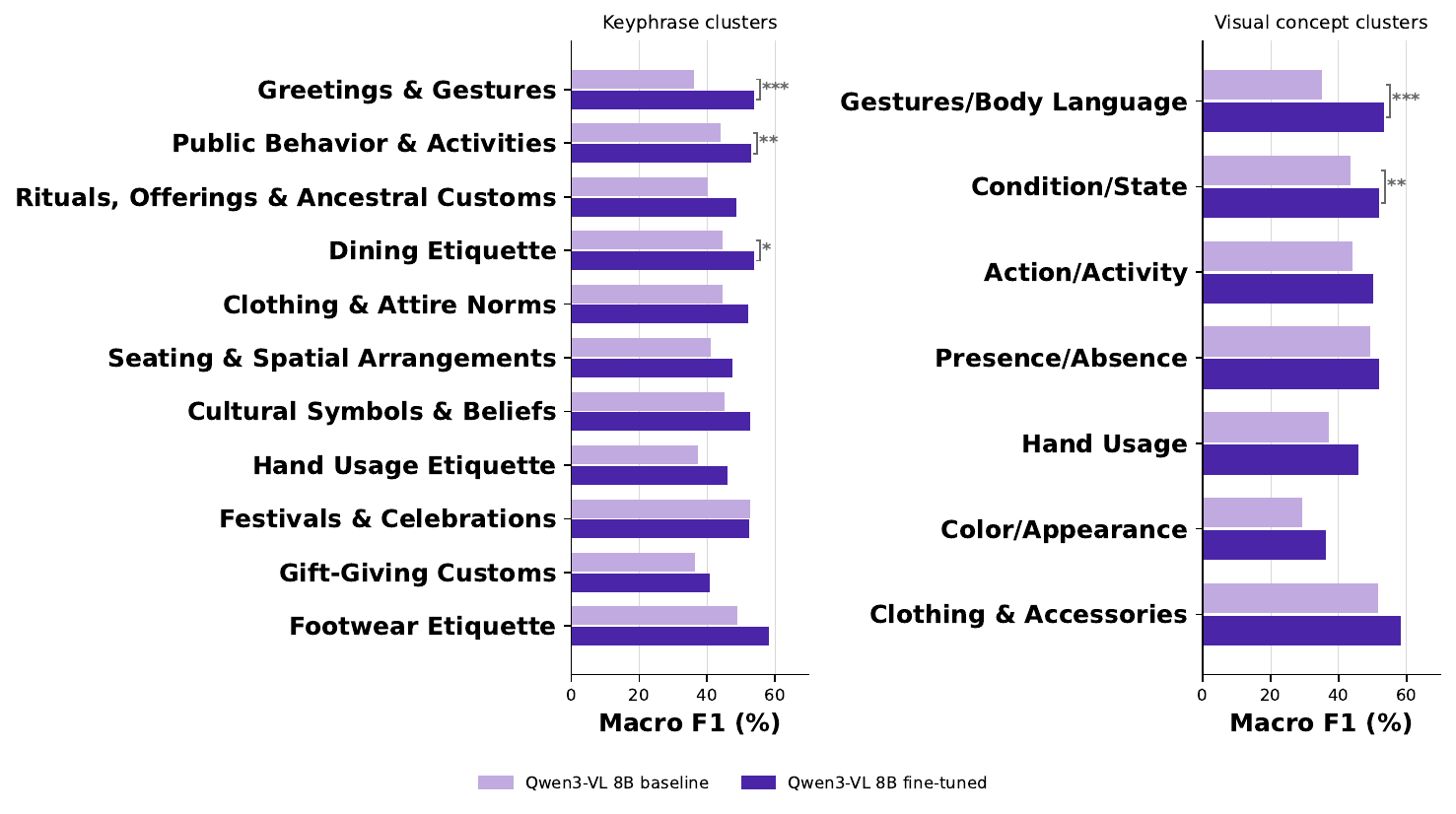}
    \caption{Improvement of finetuned Qwen3 VL 8B (follow-heavy 64k) across cultural norm domains and visual contrast types. Improvements measured using single image classification F1.}
    \label{app:fig:finetuned_8b_f1_semantic}
\end{figure}

We see significant improvement from culturally underrepresented non-Western countries, especially for the smaller finetuned Qwen3 VL 4B model. Qwen3 VL 4B model has very poor performance on \benchmarkName, and potentially less biases. Hence finetuning on \trainName significantly helps improve visual norm understanding, even surpassing finetuned Qwen3 VL 8B and Gemini 3.0 Flash. Figure \ref{app:fig:finetuned_qwen4b_across_countries}
and Figure \ref{app:fig:finetuned_qwen8b_across_countries} show performance across countries, using group accuracy and single image F1 score.
Finetuning also consistently improves cultural knowledge across several social norm domains, as well compositionality across different visual contrast types. Figures \ref{app:fig:finetuned_4b_group_acc_semantic}, \ref{app:fig:finetuned_4b_f1_semantic}, \ref{app:fig:finetuned_8b_group_acc_semantic}, \ref{app:fig:finetuned_8b_f1_semantic} show improvements for Qwen3 VL 4B and Qwen3 VL 8B models (follow-heavy 64k).

\end{document}